\documentclass[nohyperref]{article}

\usepackage{microtype}
\usepackage{graphicx}
\usepackage{subfigure}
\usepackage{booktabs} %

\usepackage{hyperref}

\usepackage[accepted]{icml2022}

\usepackage{mathtools}
\usepackage{amsthm}

\usepackage[utf8]{inputenc} %
\usepackage[T1]{fontenc}    %
\usepackage{hyperref}       %
\usepackage{url}            %
\usepackage{booktabs}       %
\usepackage{amsfonts}       %
\usepackage{nicefrac}       %
\usepackage{microtype}      %
\usepackage{xcolor}         %

\usepackage{amsfonts}
\usepackage{svg}
\usepackage{amsmath}
\usepackage{amssymb}
\usepackage{wrapfig}
\usepackage{booktabs}
\usepackage{subfigure}
\usepackage{caption}

\usepackage{enumitem}
\usepackage{tikz-cd} 
\usetikzlibrary{arrows}

\usepackage{booktabs, multirow} %
\usepackage{soul}%
\usepackage{changepage,threeparttable} %

\newcommand{\minor}[1]{{\color{black} #1}}

\newcommand{\edit}[1]{{\color{black} #1}}

\newcommand{\objlib}{\textsc{ObjLib}}  %

\newcommand{\libset}{\mathbb{L}}
\newcommand{\sceneset}{\mathbb{O}}

\newcommand{\slots}{{[K]}}

\newcommand{\fullmdp}{\mathcal{M}_\mathbb{L}}
\newcommand{\slotmdp}{\mathcal{M}_{[K]}}
\newcommand{\scenemdp}{\mathcal{M}_\mathbb{O}}
\newcommand{\pixelmdp}{\mathcal{M}_\text{pixel}}
\newcommand{\scenemdpi}{\mathcal{M}_{\mathbb{O}_i}}

\newcommand{\symn}{\Sigma_N}
\newcommand{\symk}{\Sigma_K}

\usepackage[capitalize,noabbrev]{cleveref}

\theoremstyle{plain}
\newtheorem{theorem}{Theorem}[section]
\newtheorem{proposition}[theorem]{Proposition}

\newtheorem{corollary}[theorem]{Corollary}
\theoremstyle{definition}
\newtheorem{definition}[theorem]{Definition}

\theoremstyle{remark}

\usepackage[textsize=tiny]{todonotes}

\icmltitlerunning{Toward Compositional Generalization in Object-Oriented World Modeling}

\begin{document}

\twocolumn[
\icmltitle{Toward Compositional Generalization in Object-Oriented World Modeling}

\icmlsetsymbol{equal}{*}

\begin{icmlauthorlist}
\icmlauthor{Linfeng Zhao}{nu}
\icmlauthor{Lingzhi Kong}{nu}
\icmlauthor{Robin Walters}{nu}
\icmlauthor{Lawson L.S. Wong}{nu}
\end{icmlauthorlist}

\icmlaffiliation{nu}{Khoury College of Computer Sciences, Northeastern University, MA}

\icmlcorrespondingauthor{Linfeng Zhao}{zhao.linf@northeastern.edu}

\icmlkeywords{World Modeling, Object-oriented Representations, Object-oriented Reinforcement Learning, Compositional Generalization}

\vskip 0.3in
]

\printAffiliationsAndNotice{}  %

\begin{abstract}
Compositional generalization is a critical ability in learning and decision-making.
We focus on the setting of reinforcement learning in \textit{object-oriented environments} to study compositional generalization in world modeling.
We (1) formalize the compositional generalization problem with an \textit{algebraic} approach and (2) study how a world model can achieve that.
We introduce a conceptual environment, Object Library, and two instances, and deploy a principled pipeline to measure the generalization ability.
Motivated by the formulation, we analyze several methods with \textit{exact} or \textit{no} compositional generalization ability using our framework, and design a differentiable approach, \minor{Homomorphic Object-oriented World Model (HOWM)}, that achieves \edit{soft} but more efficient compositional generalization.
\end{abstract}

\section{Introduction}

In learning and decision-making, the goal is to train models and agents that generalize to new data, tasks, and environments. 
Recently, there has been significant interest in learning transition models for object-based environments in computer vision and reinforcement learning, in particular from images~\citep{burgess2019monet,watters2019cobra,kipf2019contrastive,kossen2019structured,lin2020improving,veerapaneni2020entity,locatello2020object}. These environments are naturally factorized by their objects.

In this paper, we aim to study one widely existing form of generalization that has not been formally studied in object-oriented \edit{world modeling}: \textit{compositional generalization}.  %
That is, if we have seen various combinations of objects during training, but are presented with a novel combination at evaluation, we should still expect our agents to recognize familiar objects seen during training and predict their effects appropriately.
While compositional generalization has been previously studied in natural language~\citep{johnson2017clevr,lake2018generalization,bahdanau2018systematic,gordon2019permutation,keysers2019measuring}, such work does not yet exist for \edit{(action-conditioned) world modeling}.

\begin{figure}[t]
\centering
    \includegraphics[width=\linewidth]{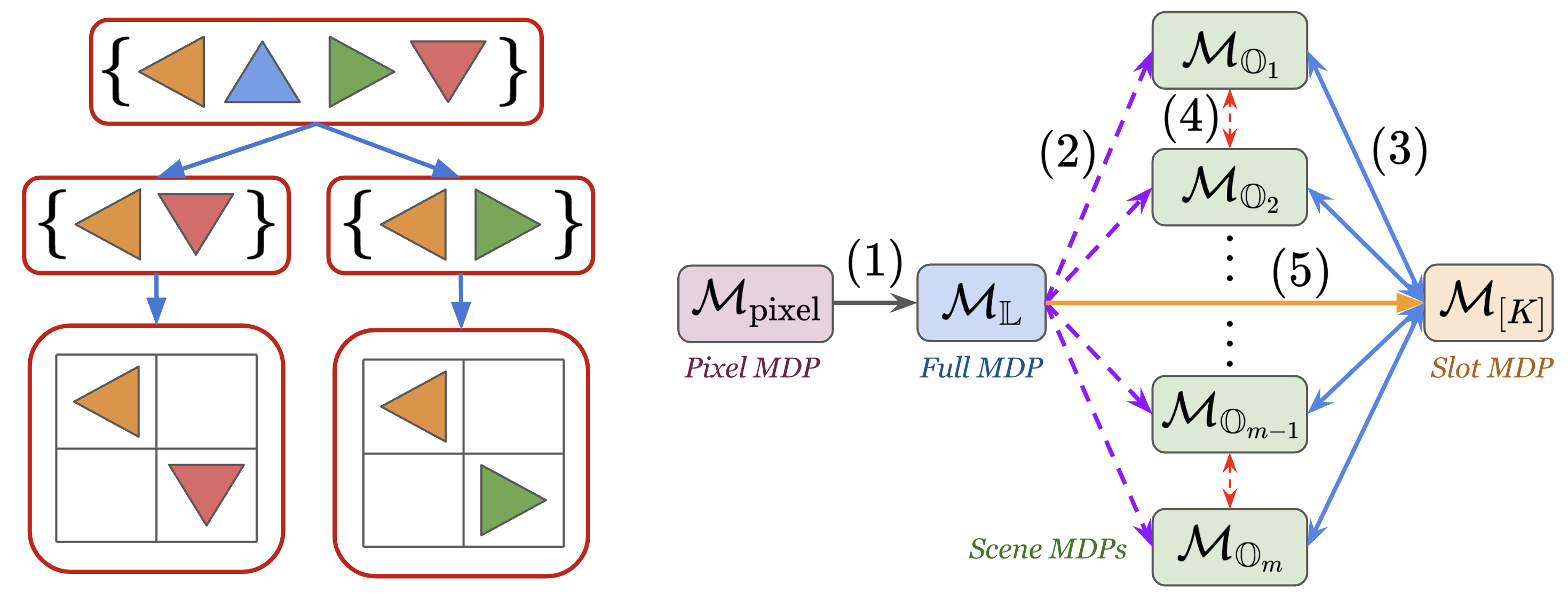}

    \caption{
    \textbf{(Left)} An example of our Object Library environment: Rush Hour, described in Section~\ref{subsec:illustrative}. Each scene features $K=2$ objects from a library of $N=4$ objects.
    \textbf{(Right)} A family of MDPs for modeling Object Library, as explained in Section~\ref{subsec:mdps}.
}
\vspace{-10pt}
\label{fig:objlib_example}  %
\end{figure}

To study compositional generalization, we design a class of object-oriented environments called Object Library, illustrated in Figure~\ref{fig:objlib_example} left. Such environments feature $K$ objects, drawn from a library of $N$ objects; the $K$ objects remain the same during each episode. Although visually distinct, these environments are isomorphic to each other (Figure~\ref{fig:objlib_example} center). Furthermore, the isomorphism is factorized by the constituent objects, so we should expect our models to generalize to new scenes (combinations of objects), as long as the constituent objects have been observed in some scenes during training. For example, if during training we have observed scenes containing $\{\textcolor{blue}{\blacktriangle},\textcolor{red}{\blacktriangledown}\}$ and $\{\textcolor{orange}{\blacktriangleleft},\textcolor{green}{\blacktriangleright}\}$, then our models should also make accurate predictions in novel scenes $\{\textcolor{blue}{\blacktriangle},\textcolor{orange}{\blacktriangleleft}
\}$ and $\{\textcolor{red}{\blacktriangledown}, \textcolor{green}{\blacktriangleright}\}$.

\edit{We formally define compositional generalization as the ability to generalize from a scene (with a combination of object) to another scene with \textit{replaced} objects, or \textit{equivariance} to \textit{object-replacement operation}.}
\edit{This bridges the behavioral perspective (generalizing to replaced object) with functional implementation (equivariant model).}
The proposed framework illustrates two paths for compositional generalization \edit{in world modeling}: (1) \textit{exact} compositional generalization with potentially more intensive resource usage, (2) \edit{learning \textit{soft}} object and action binding in \textit{latent} space.
Specifically, for the latter path, one primary challenge comes from the lack of canonical ordering of objects in images.
We prove that, if it learns to correctly \textit{bind actions} with \textit{latent object slots}, this learned path can still achieve \textit{perfect} compositional generalization, with far less resources needed.

Based on this analysis, we propose an \edit{soft} approach for learning compositional generalizable world models, called \minor{Homomorphic Object-oriented World Model (HOWM)}, which deploys an \textit{Action Attention} module to bind actions and can be trained differentiably with the \textit{Aligned Loss}.
We analyze the compositional generalization ability of existing methods and HOWM using instances of Object Library, and demonstrate that HOWM generalizes well with fewer resources.
\edit{The idea of learning action binding for compositional generalization is generic and can be \edit{plugged into} world models for more complex environments or object-based planning.}
\edit{More resources are available under \url{http://lfzhao.com/oowm}.}

\section{Background}
\label{sec:background}

Our environments are modeled as Markov decision processes. A Markov decision process (MDP) is a $5$-tuple $\mathcal{M}=\langle \mathcal{S}, \mathcal{A}, T, R, \gamma \rangle$, with state space $\mathcal{S}$, action space $\mathcal{A}$, transition function $T\colon \mathcal{S} \times \mathcal{A} \times \mathcal{S} \to \mathbb{R}_{+}$, reward function $R\colon \mathcal{S} \times \mathcal{A} \to \mathbb{R}$, and discount factor $\gamma \in [0, 1]$.

\paragraph{MDP homomorphisms.} In this paper, we will study families of related MDPs using the framework of MDP homomorphisms~\citep{ravindran2004algebraic,van2020mdp}. An \textit{MDP homomorphism} $h\colon \mathcal{M} \to \overline{\mathcal{M}}$ is a mapping from one MDP $\mathcal{M}=\langle \mathcal{S}, \mathcal{A}, T, R, \gamma \rangle$ to another $\overline{\mathcal{M}}=\langle \overline{\mathcal{S}}, \overline{\mathcal{A}}, \overline{T}, \overline{R}, \gamma \rangle$. The map $h$ consists of a tuple of surjective maps $h=\langle \phi, \{ \alpha_s \mid s \in \mathcal{S} \} \rangle$, where $\phi\colon \mathcal{S} \to \overline{\mathcal{S}}$ is the state mapping and $\alpha_s\colon \mathcal{A} \to \overline{\mathcal{A}}$ is the \textit{state-dependent} action mapping.
The mappings are constructed to satisfy the following conditions for all $s,s' \in \mathcal{S}$ and $a \in \mathcal{A}$:
\begin{equation} \label{eq:homomorphism}
\begin{aligned}
\overline{R}\left(\phi(s), \alpha_{s}(a)\right) & = R(s, a), & & 
\\
\overline{T}\left(\phi\left(s^{\prime}\right) \mid \phi(s), \alpha_{s}(a)\right) & =
\!\!\! \sum_{s^{\prime \prime} \in \phi^{-1}\left(\phi(s^{\prime})\right)} \!\!\!\!\!\! T\left(s^{\prime \prime} \mid s, a\right).
\\
\end{aligned}
\end{equation}
We call the \textit{reduced} MDP $\overline{\mathcal{M}}$ the \textit{homomorphic image} of $\mathcal{M}$ under $h$.
If $h=\langle \phi, \{ \alpha_s \mid s \in \mathcal{S} \} \rangle$ has \textit{bijective} maps $\phi$ and $\lbrace \alpha_s \rbrace$, we call $h$ an \textit{MDP isomorphism}.

\paragraph{Symmetry groups and equivarance.} In Section~\ref{sec:compgen}, we will formalize compositional generalization in object-oriented environments as a type of ``object-replacement'' symmetry. In mathematics, a \textit{symmetry group} $G$ is an algebraic concept, denoting the collection of all symmetric transformations of an entity, satisfying the axioms of associativity, identity, inverse, and closure. A (left) \textit{group operation} (action) of a group $G$ on a set $\mathcal{X}$ is defined as the mapping $(g, x) \mapsto g.x$. If $f$ is a function $f: \mathcal{X} \to \mathcal{Y}$ and $G$ acts on $\mathcal{X}$ and $\mathcal{Y}$, then $f$ is an \textit{equivariant map} if it commutes with the operation of the group: $g.f(x) = f(g.x), \forall g \in G, \forall x \in \mathcal{X}$. The function $f$ is considered \textit{$G$-equivariant}. If instead $f(x) = f(g.x)$ holds, then $f$ is considered \textit{$G$-invariant}.

\paragraph{MDP homomorphisms with symmetry.} The above concepts can be connected together. Given group $G$, an MDP homomorphism $h$ is said to be \textit{group structured} if any state-action pair $(s,a)$ and its transformed counterpart $g.(s,a)$ are mapped to the same abstract state-action pair: $(\phi(s), \alpha_{s}(a)) = (\phi(g.s), \alpha_{g.s}(g.a))$, for all $s \in \mathcal{S}, a \in \mathcal{A}, g \in G$.
For convenience, we denote $g.(s,a)$ as $(g.s,g.a)$, where $g.a$ implicitly\footnotemark{} depends on state $s$.
Applied to the transition and reward functions, the transition function $T$ is $G$-equivariant if $T$ satisfies $T(g.s' | g.s, g.a) = T(s' | s, a)$, and reward function $R$ is $G$-invariant if $R(g.s, g.a)  = R(s, a)$, for all $s \in \mathcal{S}, a \in \mathcal{A}, g \in G$.

This section is intentionally brief; see~\citet{van2020mdp} and \citet{ravindran2004algebraic} for a more in-depth account of MDPs with symmetries.
To keep the exposition simple in the main text, we focus on predicting transition dynamics only and assume that transitions are deterministic. The derivations extend naturally to stochastic environments and rewards (Appendix~\ref{sec:proofs} and~\ref{subsec:reward}).

\footnotetext{The group operation acting on action space $\mathcal{A}$ \textit{depends on state}, since $G$ actually acts on the \textit{product space} $\mathcal{S} \times \mathcal{A}$, $(g, (s,a)) \mapsto g.(s,a)$, while we denote it as $(g.s, g.a)$ for consistency with $h=\langle \phi, \{ \alpha_s \mid s \in \mathcal{S} \} \rangle$.
As a bibliographical note, in~\citet{van2020mdp}, the group acting on state and action space is denoted as state transformation $L_g\colon \mathcal{S} \to \mathcal{S}$ and \textit{state-dependent} action transformation $K_g^s\colon \mathcal{A} \to \mathcal{A}$.
}

\begin{figure*}[t]
\includegraphics[width=0.9\linewidth]{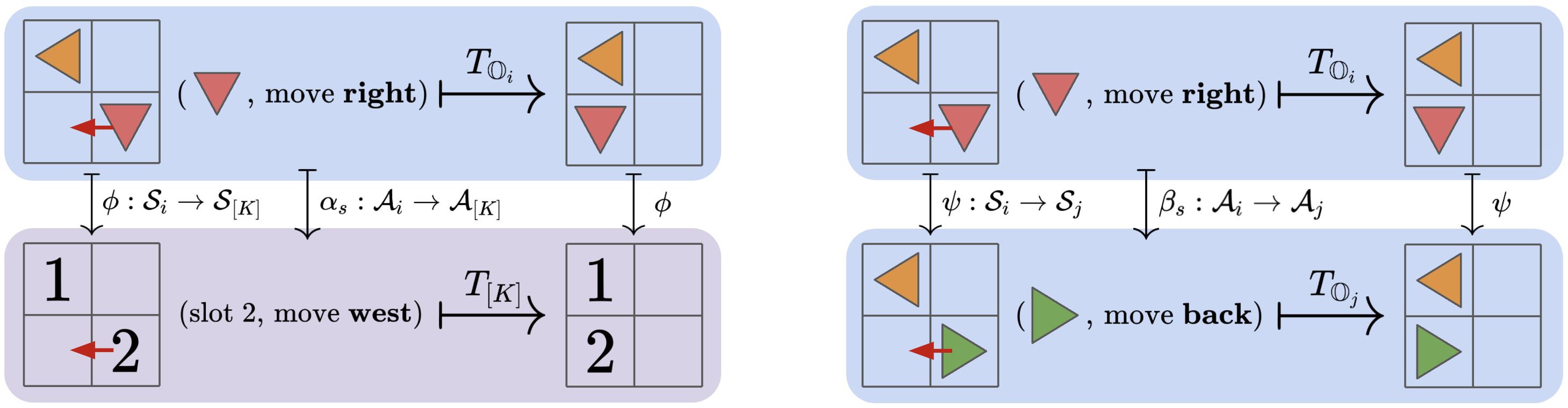}  %
\centering
\caption{
An illustrative commutative diagram.
\textbf{(Left)} For any scene MDP $\scenemdpi$, we can bind its objects and actions to the representative {slot MDP} $\slotmdp$. Note that the \textit{object-relative} action "\texttt{right}" needs to be converted to the \textit{absolute representative} action "\texttt{west}" in $\slotmdp$.
\textbf{(Right)} Since the binding between objects (and their actions) and slots is one-to-one, we can relate two {scene MDPs} by composing the binding $h_{i\to j} = h_i \circ h^{-1}_j$; the two MDPs are isomorphic.
}
\vspace{-10pt}
\label{fig:obj_commutative}  %
\end{figure*}

\section{Object-Oriented Environments}
\label{sec:ooe}

In this section, we define the concept of object-oriented environments (OOE), and introduce a specific family of OOEs, Object Library, for studying composition generalization. We model these environments as MDPs, and in the process we will define multiple related MDPs for Object Library.

\subsection{Image-based environments and factorization}
\label{subsec:ooe}

We study image-based environments with multiple objects.
At each time step, the agent observes an image $s_t \in \mathcal{S}$, consisting of multiple objects that fully describe the current \textit{scene}, and takes an action $a_t \in \mathcal{A}$ to control a single object. We model this as an MDP with image-based states, $\pixelmdp$.
What distinguishes $\pixelmdp$ from an arbitrary high-dimensional MDP with a high-dimensional state space is that the environment actually has low-dimensional factorized structure, in that the environment is fully determined solely by the objects in the scene and their relations~\citep{ravindran2004algebraic,diuk2008object}.

More formally, we define $\pixelmdp$ to be an object-oriented environment (OOE) if there exists an MDP homomorphism $\pixelmdp \to \overline{\mathcal{M}}$ that maps the image-based MDP to a factored MDP $\overline{\mathcal{M}}$~\citep{guestrin2003efficient}. $\overline{\mathcal{M}}$ has factorized state and action spaces $\mathcal{\overline{S}} = \mathcal{\overline{S}}_1 \times \ldots \times \mathcal{\overline{S}}_N, \mathcal{\overline{A}} = \mathcal{\overline{A}}_1 \times \ldots \times \mathcal{\overline{A}}_N$, where each factor is meant to model one of $N$ objects, and ideally has significantly lower dimension.
We explain why we choose factorized action space in \minor{Appendix \ref{sec:action_space}}.

\subsection{Object Library}
\label{subsec:objlib}

In this paper, we consider compositional generalization within a family of OOEs, which we call \textit{Object Library}. Object Library is an OOE augmented with a library of all possible objects $\libset = \{ o_1, \ldots, o_N \}$, of which only $K$ objects are present in any given scene, where $1 < K < N$. Note that the library is not given and is meant to be learned from images; we merely introduce it as a conceptual tool.

In each episode, the $K$ objects are persistent, but between episodes a different set of $K$ objects may be chosen from $\libset$. This allows us to generate different scenes with different object sets during training and evaluation, thus enabling us to measure the performance discrepancy when faced with different \emph{object compositions}. Compositional generalization will be formally defined in the next section.

\subsection{Modeling Object Library as a family of MDPs}
\label{subsec:mdps}

We now define several MDPs for modeling Object Library, illustrated in Figure~\ref{fig:objlib_example}.
\begin{itemize}[leftmargin=*]
    \item Pixel MDP $\pixelmdp$: This is the original environment with image-based states.
    \item Full MDP $\fullmdp$: We define $\fullmdp$ to be the latent factored MDP, with one state/action factor per object: $\mathcal{S}_\libset = \mathcal{S}_1 \times \ldots \times \mathcal{S}_N$, and likewise for $\mathcal{A}_\libset = \mathcal{A}_1 \times \ldots \times \mathcal{A}_N$. Since there are only $K$ objects in the scene, each $\mathcal{S}_i$ contains a special null state $\varepsilon$ that indicates the object is not present.  %
    \item Scene MDP $\mathcal{M}_\sceneset$: Since the $K$ objects persist in any particular instance of the Object Library environment, both $\pixelmdp$ and $\fullmdp$ are partitioned into $N \choose K$ connected components, or \textit{sub-MDPs}~\citep{ravindran2004algebraic}. It is convenient to consider each of these sub-MDPs, which we call the \textit{scene MDP} $\mathcal{M}_\sceneset$. More formally, let $\sceneset = \{ i_1, \ldots, i_K \}$ be the indices of the $K$ present objects. We construct $\mathcal{M}_\sceneset$ by projecting $\mathcal{S}_\libset$ and $\mathcal{A}_\libset$ to $\sceneset$: $\mathcal{S}_\sceneset = \mathcal{S}_{i_1} \times \ldots \times \mathcal{S}_{i_K}$, and likewise for $\mathcal{A}_\sceneset$.
    \item Slot MDP $\slotmdp$: We will eventually show that all scene MDPs are isomorphic to each other. Key to this construction is the existence of a \emph{representative} factored MDP among all scenes, which we denote as the \textit{slot MDP} $\slotmdp$. $\slotmdp$ has $K$ object ``slots'': $\mathcal{S}_\slots = \mathcal{S}_1 \times \ldots \mathcal{S}_K$, and likewise for $\mathcal{A}_\slots$. Ideally, we can "bind" every scene MDP $\scenemdp$ to $\slotmdp$ \textit{without} any loss of information, since a scene MDP has $K$ objects and the slot MDP can represent the $K$ objects using its $K$ slots.
\end{itemize}

The relationship between these MDPs are shown in Figure~\ref{fig:objlib_example} right. There exists an MDP homomorphism from the high-dimensional $\pixelmdp$ to the low-dimensional factored $\fullmdp$ (black arrow). Depending on the $K$ objects present in the scene, $\fullmdp$ can be projected onto one of $N \choose K$ scene MDPs $\scenemdpi$ (purple dashed arrow). Each of these $\scenemdpi$ is isomorphic to $\slotmdp$ (blue arrow, double headed), and therefore are also isomorphic to each other (red dashed arrow, double headed). Our analysis in Section~\ref{sec:compgen} also assumes the existence of an MDP homomorphism from $\fullmdp$ to $\slotmdp$ (orange arrow).

\section{Compositional Generalization in OOEs}
\label{sec:compgen}

In OOEs, at least two types of generalization exist: (1) generalizing to \textit{unseen objects}, and (2) generalizing to \textit{unseen combinations} (scenes) of \textit{known objects}. We consider type (2) as \textit{compositional generalization}. This is similar to recent work on compositional generalization in natural language~\citep{gordon2019permutation}, where a sentence is viewed as an ordered set of words. In the context of Object Library, the object library $\libset$ is our ``vocabulary'', and selecting novel sets of objects $\sceneset$ to form scenes is similar to composing unseen sentences from novel combinations of words.

We first illustrate this key idea using a simple instance of Object Library. Next, we formally define our notion of compositional generalization. Finally, we provide a set of conditions that we prove are sufficient to achieve compositional generalization in the Object Library family of OOEs.

\subsection{An illustrative example}
\label{subsec:illustrative}

We use an instance of Object Library, Rush Hour, to illustrate our ideas (see Figures~\ref{fig:objlib_example} and~\ref{fig:obj_commutative}).
Rush Hour is motivated by a grid game, where multiple cars are in different absolute orientations: \texttt{(north, south, east, west)}.
Each car has an action space \texttt{(forward, backward, left, right)} that moves it relative to its orientation.
For example, if a car is facing \texttt{south}, moving \texttt{right} means moving \texttt{west} in the world.
Consider an instance with $N=4$ possible objects in the library $\libset$ described by shape and orientation: \{\textcolor{blue}{$\blacktriangle$},
\textcolor{red}{$\blacktriangledown$},
\textcolor{orange}{$\blacktriangleleft$},
\textcolor{green}{$\blacktriangleright$}\}.
Each scene consists of $K=2$ objects, so there are $6$ possible scenes: \{
\{\textcolor{blue}{$\blacktriangle$},
\textcolor{red}{$\blacktriangledown$}\},
\{\textcolor{blue}{$\blacktriangle$},
\textcolor{orange}{$\blacktriangleleft$}\},
\{\textcolor{blue}{$\blacktriangle$},
\textcolor{green}{$\blacktriangleright$}\},
\{\textcolor{red}{$\blacktriangledown$},
\textcolor{orange}{$\blacktriangleleft$}\},
\{\textcolor{red}{$\blacktriangledown$},
\textcolor{green}{$\blacktriangleright$}\},
\{\textcolor{orange}{$\blacktriangleleft$},
\textcolor{green}{$\blacktriangleright$}\}
\}.

\textbf{Object-replacement symmetry.}
We formalize the notion of mapping between scenes (with different object combinations) using permutation symmetry. The symmetry group $\symn=\Sigma_4$ acts naturally on $\mathbb{L}$ with $4$ cars, by replacing a car with another; it induces an operation acting on scenes. For example, if we have a permutation mapping $\sigma(\textcolor{red}{\blacktriangledown})=\textcolor{green}{\blacktriangleright}, \sigma \in \Sigma_4$, and other cars remain the same, the induced operation on the scene would be $\sigma(\{\textcolor{orange}{\blacktriangleleft},\textcolor{red}{\blacktriangledown}\})= \{\textcolor{orange}{\blacktriangleleft},\textcolor{green}{\blacktriangleright}\}$.
If our learned transition model is equivariant to $\symn$, then we can generalize to novel scenes (e.g., Figure~\ref{fig:obj_commutative} bottom right) by appropriate permuting (replacing objects) from training scenes (Figure~\ref{fig:obj_commutative} top right).

\textbf{Lifting from the slot MDP.}
We can observe that scenes with $K$ different present objects are structurally similar to each other, motivating us to not consider all $N$ objects independently, but to share knowledge between scenes.
Furthermore, since all $K$-object scenes are similar, we could find a \textit{representative} ``scene'' where all scenes are similar to it, and convert actions to consistent meaning (from relative \texttt{up} to absolute \texttt{north}, Figure~\ref{fig:obj_commutative} bottom left).
We call it the \textit{slot MDP} $\slotmdp$ (Figure~\ref{fig:obj_commutative} bottom left), where all scene MDPs are called \textit{isomorphic} to it.
In other words, by learning a world model in the slot MDP, we should automatically get good model for all scenes, or ``lifting'', and thus generalize.

\textbf{Measuring compositional generalization.}
If we train on scene MDPs \{$\mathcal{M}_{\{\textcolor{blue}{\blacktriangle},
\textcolor{red}{\blacktriangledown}\}}, \mathcal{M}_{\{\textcolor{orange}{\blacktriangleleft},
\textcolor{green}{\blacktriangleright}\}} $\}, the model should ideally generalize to
\{$\mathcal{M}_{\{\textcolor{blue}{\blacktriangle},\textcolor{orange}{\blacktriangleleft}
\}}, \mathcal{M}_{\{\textcolor{red}{\blacktriangledown},
\textcolor{green}{\blacktriangleright}\}} $\}.
By training on the former and evaluating transition-function prediction errors on the latter, we can measure generalization error. We formalize this by defining the notion of \textit{equivariance error} in the next section.
See Section~\ref{sec:measuring-cg} for details.

\begin{figure*}[t]
\centering
\subfigure{
    \includegraphics[width=0.62\linewidth]{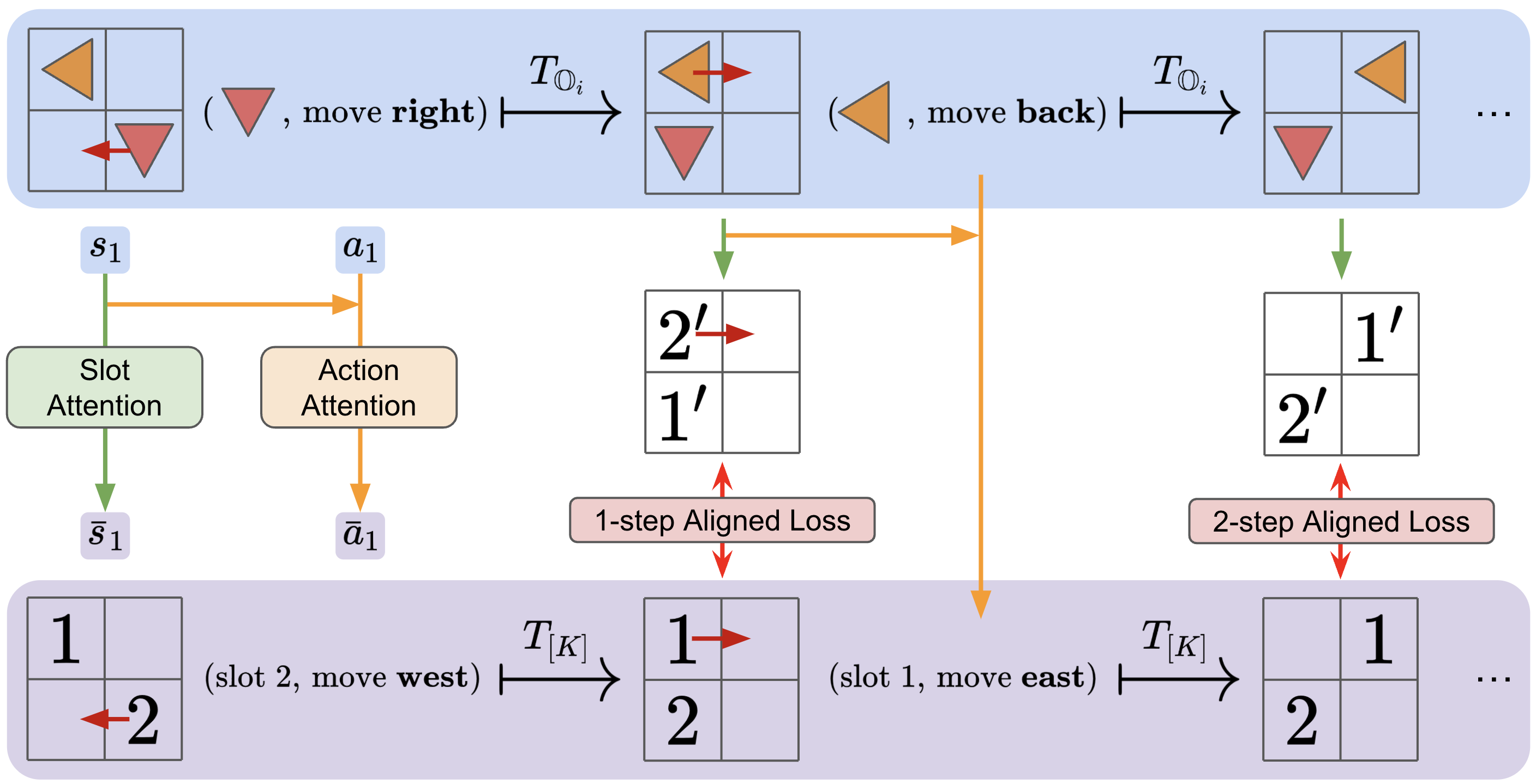}
}
\hfill
\subfigure{
    \includegraphics[width=0.33\linewidth]{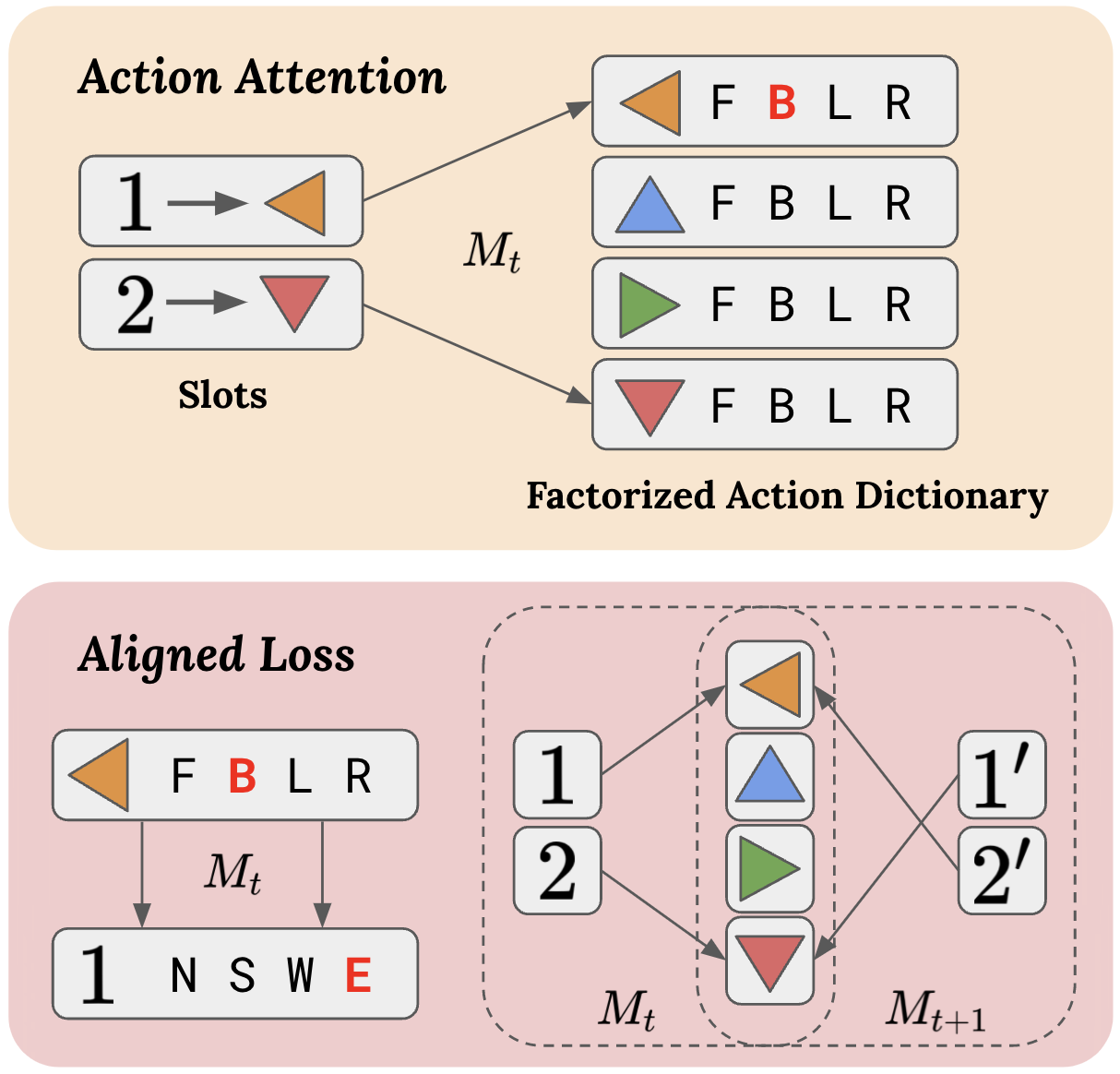}
} 
\vspace{-10pt}
\centering
\caption{
\textbf{(Left)} Overview of the world model prediction: the upper blue sequence is a ground pixel MDP with some objects $\mathbb{O}_i$, and the lower one is the slot MDP.
We emphasize two facts: (1) encoded object slots in different steps may have different ordering (marked as $1'$ and $2'$), and (2) the transition model is equivariant in slot ordering, i.e., consistent across time steps (in $1$ and $2$), thus the loss computation needs alignment of slots (between $1,2$ and $1',2'$).
\textbf{(Top right)} \textcolor{orange}{Action Attention} learns to bind actions from interaction (action-object correspondence is \textit{unknown} and learned).
\textbf{(Bottom right)} In the \textcolor{red}{Aligned Loss}, the learned binding matrices $M_t$ and $M_{t+1}$ are used to lift slots in $t$ and $t+1$ to a canonical space (full MDP).
}
\vspace{-10pt}
\label{fig:method_overview}
\end{figure*}

\subsection{Defining exact compositional generalization} %
\label{subsec:definition}

Our definition is based on equivariance to object replacement in OOEs. Specifically, if the (learned) transition model $\hat{T}_\libset$ for $\fullmdp$ is equivariant to any object-set replacement, then $\hat{T}_\libset$ can generalize to all object sets $\sceneset$, including unseen combinations, as long as all individual objects in the library $\libset$ have been observed in training scenes. 
\edit{This equivariance definition connects the behavioral perspective (how a system with compositional generalization should behave) and functional perspective (achieving by permutation equivariant networks).}
More formally, we define the object-replacement operation as follows:

\begin{definition}[Object-replacement operation $p_\libset$]
\label{def:obj_rep_op}
Let $\symn$ denote the permutations of $N$ elements, which acts naturally on the object-library set $\libset$.
The object-replacement operation is the group operation $\rho_\libset\colon \symn \times \libset \to \libset$ where $\symn$ acts on the indices in $\libset$ to replace the identity of objects.
\end{definition}

For example, using disjoint cycle notation, an object replacement operation $\pi = (1 2 3)(45)$ acts on $\libset$ by sending $\pi(3) = 1$ and $\pi(4) = 5$. This induces a group operation on scenes, $p_K$ (see Definition~\ref{def:scene_op} in Appendix~\ref{subsec:induced_operations}), that maps a set of objects to another set, e.g., $p_K(\pi,\{ 3, 4 \}) = \{ 1, 5 \}$. Thus, $\Sigma_N$ permutes the scene MDPs $\lbrace \mathcal{M}_\sceneset \rbrace$.
We use $p_\libset$ to define the measure of compositional generalization:
\begin{definition}[Equivariance error of $\hat{T}_\libset$ on $\fullmdp$]
\label{def:equivariance-error}
Let $\hat{T}_\libset: \mathcal{S}_\libset \times \mathcal{A}_\libset \times \mathcal{S}_\libset \to \mathbb{R}_{+}$ be a (learned) transition model of $\fullmdp$.
The sample equivariance error of $T_\libset$ at $(s,a,s') \in \mathcal{S}_\libset \times \mathcal{A}_\libset \times \mathcal{S}_\libset$ with respect to $\sigma \in \Sigma_N$ is defined
\begin{equation}
\lambda_{\libset}^\sigma \triangleq \left| 
\hat{T}_\libset(s' \mid s,a) - \hat{T}_\libset(\sigma.s' \mid \sigma.s, \sigma.a) 
\right|.
\end{equation}
The equivariance error of $T_\libset$ is then defined as the expectation $\lambda_\libset = \mathbb{E}_{s,a,s',\sigma} [\lambda_\libset^\sigma]$.
\end{definition}

The magnitude of $\lambda$ measures the failure of $\hat{T}_\libset$ to be $\Sigma_N$-equivariant, hence the name \textit{equivariance error}. For a perfectly $\symn$-equivariant transition function $T_\libset$, the error is guaranteed to be $0$ by the definition of equivariance.
Note that for clarity, we only consider \textit{transition} prediction in the above definition. A complete version of homomorphism also preserves the reward structure, thus preserving optimal values and policies. We provide the full version with rewards in Appendix~\ref{subsec:reward}.

\subsection{\edit{Achieving soft} compositional generalization}
\label{subsec:lifting}

Ideally, we can achieve compositional generalization according to the above metric by simply learning a $\symn$-equivariant transition model and correct binding of objects \textit{and} actions.
However, this is difficult in practice for large $N$ (size of object library), as we show in our experiments. In this section, we derive an alternative path to $\symn$-equivariance that only requires us to learn a $\symk$-equivariant model, which is significantly more tractable since we expect $K \ll N$.

Instead of directly learning a $\symn$-equivariant model, which is difficult for large $N$, we can instead learn a $\symk$-equivariant model for $\slotmdp$, and ``lift'' it to achieve $\symn$-equivariance. 
We need to first define what is ``good'' binding, which intuitively means that we can project $N$ objects to a subspace of $K$ slots, while we are still able to identify them, i.e., how $K$-slot identities are permuted.

\begin{definition}[Projection property]
Let $h$ be an MDP homomorphism from the full MDP to the slot MDP, $h: \fullmdp \to \slotmdp$.
The homomorphism $h = \langle \phi, \{ \alpha_s \mid s \in \mathcal{S}  \} \rangle$ satisfies the projection property if, for any $\sigma \in \symn$, $s \in \mathcal{S}_\libset$, $a \in \mathcal{A}_\libset$, there exists $\overline{\sigma} \in \symk$ such that
\begin{equation}
    \overline{\sigma}.\phi(s) = \phi(\sigma.s) \quad \text{ and } \quad \overline{\sigma}.\alpha_s(a) = \alpha_{\sigma.s}(\sigma.a) \;.
\end{equation}
\end{definition}
In other words, the $K$ present objects in $s \in \mathcal{S}_\libset$ \edit{are \textit{bound}} to the $K$ slots in $\phi(s) \in \mathcal{S}_{[K]}$ in a \textit{specific} order.
It implicitly assumes the binding of present objects in $\fullmdp$ to slots in $\slotmdp$ is in some \textit{temporally consistent} order: $\mathcal{S}_{i_k} \mapsto \mathcal{S}_k$.

By definition of MDP homomorphisms, given a model $\hat{T}_\libset$ for $\fullmdp$, $h$ induces a model $\hat{T}_\slots$ for $\slotmdp$. If $\hat{T}_\libset$ has equivariance error $\lambda_\libset$, we can show that $\hat{T}_\slots$ will have the following equivariance error \footnotemark{}:
\begin{proposition} \label{prop:error}
Let $h: \fullmdp \to \slotmdp$ be an MDP homomorphism satisfying the projection property. Suppose $\hat{T}_\libset$ is a (learned) transition model of $\fullmdp$ with equivariance error $\lambda_\libset$. Then $\hat{T}_\slots$, the induced transition model of $\slotmdp$ under $h = \langle \phi, \{ \alpha_s \mid s \in \mathcal{S} \} \rangle$, has \textit{sample} equivariance error at $(\phi(s),\alpha_{s}(a),\phi(s')) \in \mathcal{S}_\slots \times \mathcal{A}_\slots \times \mathcal{S}_\slots$ and $\overline{\sigma} \in \symk$:
\begin{align}
& 
\lambda^{\overline{\sigma}}_\slots \triangleq
\left[
\left| 
\hat{T}_\slots(\phi\left(s^{\prime}\right) \mid \phi(s), 
\alpha_{s}(a))  
-  \nonumber \right. \right. \\
&  \ \ \ \  \left. \left. \hat{T}_\slots(\overline{\sigma}. \phi\left(s^{\prime}\right) \mid \overline{\sigma}. \phi(s), \overline{\sigma}. \alpha_{s}(a)) 
\right|
\right]
= C \cdot \lambda_{\libset}^\sigma \;,
\end{align}
where $C = {N \choose K}$ is the number of $K$-slot scenes given an $N$-object library, $\phi: \mathcal{S}_\libset \to \mathcal{S}_\slots$ and $\alpha_s: \mathcal{A}_\libset \to \mathcal{A}_\slots$.
The equivariance error is then $\lambda_\slots = \mathbb{E}_{s,a,s',\overline{\sigma}} [\lambda_\slots^{\overline{\sigma}}] = C \cdot \lambda_\libset$.
\end{proposition}

\footnotetext{However, from the computational perspective, the information of binding to latent space is unknown, so the numerical values are hard to compute, as detailed in Appendix~\ref{sec:measuring-cg}.}

The proof is provided in Appendix~\ref{sec:proofs}. Therefore, if $\hat{T}_\libset$ has perfect compositional generalization (equivariance error $\lambda_\libset = 0$), and homomorphism $h$ satisfies the projection property, then the induced model $\hat{T}_\slots$ is $\symk$-equivariant.
Conversely, since the proposition holds with equality, if we have a $\symk$-equivariant model $\hat{T}_\slots$ in $\slotmdp$, and $\slotmdp$ is a homomorphic image of $\fullmdp$, then we can lift the model to a $\symn$-equivariant model $\hat{T}_\libset$ in $\fullmdp$, which allows us to simplify $\symn$-equivariant models.

\begin{corollary} [Lifted model is $\symn$-equivariant]
If (1) $\hat{T}_\slots$ is $\symk$-equivariant, and (2) there exists a homomorphism $h: \fullmdp \to \slotmdp$ satisfying the projection property, then $\hat{T}_\libset$ is $\symn$-equivariant.
\end{corollary}

\subsection{\edit{Practically measuring compositional generalization}}

\edit{
Even though we formally define the compositional generalization error as an expectation $\mathbb{E}_{s,a,s',\overline{\sigma}} [\lambda_\slots^{\overline{\sigma}}]$ over all transition tuples and all permutations, directly computing it in \textit{learned latent space} is not practical.
The reason is that we only assume there exists correct object and action binding $\phi,\alpha_s$ (by the projection property), while they are \textit{learned} by models and \textit{not} necessarily correct.

In the definition, several sources of error exist: (1) \textit{prediction error} $(s, a) \to s'$, (2) \textit{binding error} $(\phi(s), \alpha_s(a) ) \to \phi(s')$, (3) \textit{compositional error} $(\sigma.\phi(s), \sigma.\alpha_s(a) ) \to \sigma.\phi(s')$.
Alternatively, we need a practical metric to bypass (2) and measure (1)+(3).
To this end, motivated by generating compositional natural language data \citep{keysers2019measuring}, we propose to train and test on disjoint set of scenes, and directly measure the prediction error on test scenes, which avoids explicitly "replacing" object in latent space $\sigma.\phi(s)$.

Specifically, we need to follow two rules analogously:
(1) \textit{Disjoint scene object set.} Training and test set of object scenes should be \textit{disjoint}: $|\sceneset| = K$, $\{ \sceneset_\text{train} \} \cap \{ \sceneset_\text{test} \} = \emptyset$.
(2) \textit{Similar object distribution: $i_n \in \libset$.} Training and test datasets should have similar object frequency (uniform), to ensure the error is not biased by errors in object detection. The training set should also contain \textit{all} objects in library $\bigcup \sceneset_\text{train} = \libset$.
In the experiment section, we then use this strategy to collect data and measure prediction error on test set as a proxy of compositional generalization error.
}

\section{Compositionally Generalizable WMs}
\label{sec:worldmodel}

In the previous section, we provided the mathematical foundations for achieving compositional generalization via $\symn$- and $\symk$-equivariant transition models in $\fullmdp$ and $\slotmdp$ respectively.
Now, we provide a practical approach, based on the construction in Section~\ref{subsec:lifting}, for achieving soft $\symn$-equivariance in only $\symk$ latent space.
The overview of the model is provided in Figure~\ref{fig:method_overview}.

We focus on \textit{end-to-end} learning a world model in latent space.
The key challenge is that the object binding is unknown and no canonical ordering can be determined purely from images, so the model needs to infer it from data.
We assume the action space is factorized by objects \citep{guestrin2003efficient,kipf2019contrastive}; further explained in Appendix~\ref{sec:action_space}.
While the factorization $\mathcal{A}_\libset = \mathcal{A}_1 \times \ldots \times \mathcal{A}_N$ is fixed, the model must use interaction data $(s, a, s')$ to \emph{infer} which action controls which object.

\textbf{Stage 1: Object Extraction.}
Object extraction, or object binding~\cite{greff2020binding}, learns object-structured representations, where each present object is represented by a latent vector $\mathbb{R}^{D}$, or \textit{slot}.
All slots form a latent state $s_t$, where no canonical order exists and the order differs in different time steps.
This stage implicitly requires that \textit{all} objects in $\libset$ must have been observed during \textit{training};
at generalization time we assume it can provide representations for any individual object.
We use Slot Attention~\citep{locatello2020object} trained with reconstruction loss.

\textbf{Stage 2: Action Binding.}
In learning the transition model $T(s, a) = s'$, we need to correctly align factorized actions $a$ and object slots $s$, i.e., understand which action is controlling the object in each slot.
This corresponds to the projection property required by the homomorphism $h: \fullmdp \to \slotmdp$.
We design an attention module, named \textbf{Action Attention}, that learns this purely from interaction $(s, a, s')$, as shown in Figure \ref{fig:method_overview} (top right).

The attention matrix $M \in \mathbb{R}^{K \times N}$ is a (jointly learned) binding matrix from slots\footnotemark{} $s_t$ (at each time step) to some object identity, represented as an identity matrix $\mathrm{Id}_{N \times N}$ in our case.
This implements the \textit{state-dependent} action transformation $\alpha_s(a)$, by \textit{binding $K$ actions} corresponding to the object slots: $\alpha_{s_t}(a_t)=M_t(s_t) a_t = \bar{a}_t$, and $M_t$ is:
\begin{equation}
M_t (s_t) = \underset{K}{\operatorname{softmax}}\left(\frac{1}{\sqrt{D}} k\left( \mathrm{Id}_{N \times N} \right) \cdot q\left( s_t \right)^{T}\right) \;,
\end{equation}
where $k, q$ are linear projections mapping to some dimension $D$, and $a_t$ acts as the value $v$ without further transformation.
Intuitively, the object identity specifies a “name” to factorized actions and can be viewed as a preference for some canonical ordering of objects in the full MDP $\fullmdp$.

\footnotetext{In practice, we use $K+1$ slots to also model the background, as in Slot Attention \citep{locatello2020object}.}

\textbf{Stage 3: $\symk$-equivariant Transition Modeling.}
Proposition \ref{prop:error} implies that we only need a $\Sigma_K$-equivariant transition function for modeling the slot MDP (instead of $\Sigma_N$ for the full MDP), hence we use a message-passing GNN \citep{kipf2019contrastive}.
Crucially, we only require $K \ll N$ nodes, and since this leads to message-passing complexity of $O(K^2)$ instead of $O(N^2)$, we expect our model to be significantly more efficient.
We also assume that interactions between any objects (slots) can be modeled by a one-edge network. This can be viewed as an OO-MDP \citep{diuk2008object} with only one object class.

\textbf{Training: Aligned Loss.}
The binding matrix $M_t \in \mathbb{R}^{K \times N}$ in Action Binding relates slots at time $t$ to specific object identities.
Thus, we use it to \textit{align} slots $s_t$ and $s_{t+1}$ %
to the canonical full MDP space as $s_t^\uparrow$ and $s_{t+1}^\uparrow$ (Figure~\ref{fig:method_overview} (bottom right)) using the pseudoinverse: $s_t^\uparrow =  M_t^+ \bar{s}_t$, $s_{t+1}^\uparrow = M_{t+1}^+\bar{s}_{t+1}$, where $M_t^+ = (M_t M_t^\top)^{-1}M_t^\top \in \mathbb{R}^{N \times K}$, and $M_t^+ T(\bar{s}_t, \bar{a}_t) \approx M_{t+1}^+\bar{s}_{t+1}$.
We \emph{jointly} train the model with the \textit{aligned} loss between $K$ slots, using the object-structured contrastive loss~\citep{kipf2019contrastive}; the positive part is:
\begin{equation}
\mathcal{L}^+(s_{t}^\uparrow, s_{t+1}^\uparrow) = \left\| \mathtt{NG}(M_{{t+1}}^+) \bar{s}_{{t+1}} - \mathtt{NG}(M_{{t}}^+) T(\bar{s}_t, \bar{a}_t) \right\|^2 \;,
\end{equation}
where $\mathtt{NG}(\cdot)$ is \textit{no-gradient} operation to prevent the shortcut to change the binding matrix $M_t$ outside Action Attention.
This forms a differentiable loss and allows us to train the transition model in latent space, which we call \textbf{Homomorphic Object-oriented World Model} (HOWM).

\edit{\textbf{Inference: Multi-step prediction in latent space.}}
\edit{
For multi-step prediction at inference for evaluation, illustrated in Figure~\ref{fig:method_overview}, all slots need to be aligned with their corresponding actions with \textit{learned} binding $M_t$, since their ordering is potentially different at all time steps.
If aligning slots sequentially with learned $M$ as $T( T(\bar{s}_t, \bar{a}_t), M_{t} M_{t+1}^+ \bar{a}_{t+1} ) \approx M_{t} M_{t+2}^+ \bar{s}_{t+2}$ and compute difference, it would suffer from compounding error.
Thus, we instead align \textit{actions}: $T(s_t, M^+_{t+1}M_t {a_t}) = \hat{s}_{t+1}$, which gives preceding slots $\hat{s}_{t+1}$ (or telescoping for $t+k$) in the same order as $s_t$.
}
More details in the Appendix~\ref{sec:howm_details}.

\section{Experiments}

We study how compositional generalization is achieved in practice, by using Object Library and generalization metrics.
We provide additional results in \minor{Appendix \ref{sec:additional_results}}.

\begin{table*}[t]
\centering
\caption{
\minor{Results for all methods on the Shapes environment with $K=5$ and $N=5,10,20,30$ (four numbers in each cell). ``OM'' stands for out of GPU memory, where we limit the usage to 10GB.
We report for memory usage on $N=20$.
}
}
\vspace{-5pt}
\label{tab:shapes}
\scriptsize
\begin{tabular}{ll|ll|lll}\toprule\textbf{CG Type} & \textbf{Env=Shapes} & \textbf{Eval MRR} (\%, 1-step) & \textbf{Eval MRR} (\%, 5-step) & \textbf{Train MRR} (\%, 5-step) & \textbf{Gap}  (MRR \%, 5-step) & Memory  \\
\midrule
\multirow{1}{*}{(1. Exact CG)} 
&  $\Sigma_N\text{-CSWM}$  &100.\; 100.\; 99.9\; \textit{OM} & 99.9\; 99.9\; 99.9\; \textit{OM} & 100.\; 100.\; 100.\; \textit{OM} & \; 0.0\; \; 0.0\; \; 0.1\; \textit{OM} & 8.1GB \\

\midrule
\multirow{4}{*}{(2. No guaranteed CG)} 
  &  $\Sigma_K\text{-CSWM}$  & 100.\; 80.4\; 70.8\; 74.1 & 99.9\; 43.7\; 32.2\; 29.4 & 100.\; 100.\; 100.\; 100. & \; 0.1\; 55.3\; 67.8\; 70.6 & 1.5GB \\
&  $\Sigma_K\text{-CSWM(CA)}$  & 95.0\; 72.0\; 71.4\; 80.9 & 85.0\; 26.8\; 22.7\; 24.3 & 96.3\; 96.5\; 96.1\; 98.0 & 11.3\; 69.7\; 73.4\; 73.7 & 1.6GB   \\
&  $\text{C-WM(N)}$  & 83.6 \; 81.4\; 25.8\; 12.9  & 61.8\; 55.0\; 11.2\; 7.6 & 88.0\; 96.6\; 41.5\; 33.9 & 26.2\; 41.6\; 30.3\; 26.2 & 1.3GB \\
&  $\text{MONet(N)+BM}$ & 12.6\; 73.9\; 35.9\; \textit{OM} & \; 2.0\; 20.2\; 55.9\; \textit{OM} & \; 7.0\; 64.5\; 84.8\; \textit{OM} & \; 5.0\; 44.3\; 29.0\; \textit{OM} & 9.3GB \\
\midrule
\multirow{1}{*}{(3. \edit{Soft} CG)} 
&  $\textbf{HOWM}$ \textbf{(ours)}  & 99.2\; 98.5\; 99.7\; 99.7 & 92.3\; 84.2\; 75.1\; 81.8 &97.0\; 96.9\; 98.0\; 98.1 & \; 4.7\; 12.7\; 22.9\; 16.3 & 3.7GB \\
\bottomrule
\end{tabular}
\vspace{-10pt}
\end{table*}

\subsection{Experimental Setup}

\begin{figure}[t]
    \subfigure{
        \includegraphics[width=0.22\textwidth]{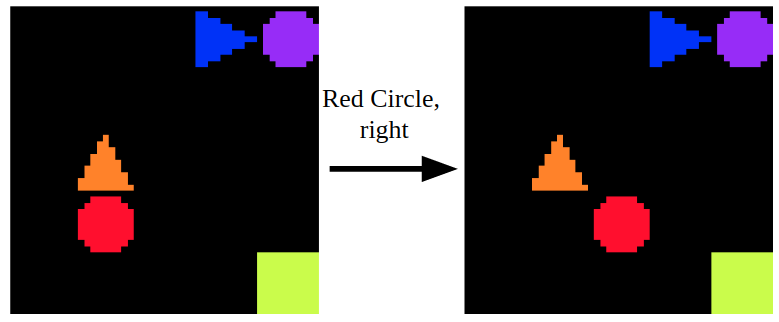} 
    }
    \hfill
    \subfigure{
        \includegraphics[width=0.22\textwidth]{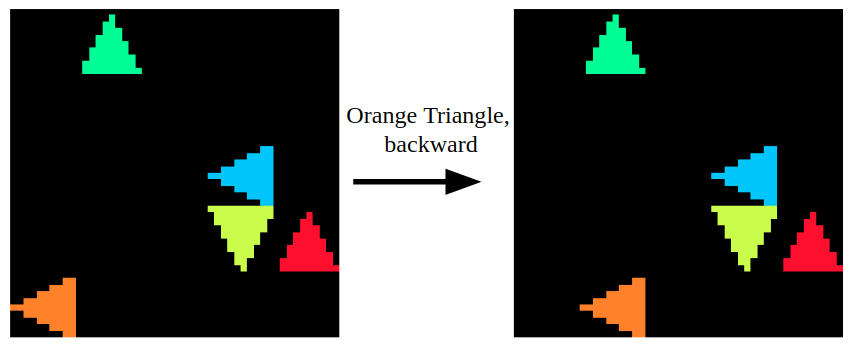}
    }
    \caption{
    Example transitions of Shapes (left pair) and Rush Hour environment (right pair).
    }
    \vspace{-10pt}
    \label{fig:obj_lib_demo} 
\end{figure}

\paragraph{Environments.}

We designed two instances of the Object Library, \objlib{}, environment. %
They are built upon the $2$-D shape version of the Block Pushing environment \cite{kipf2019contrastive}.
(1) Basic \textbf{Shapes}. In the basic case, we equip the environment with a library of objects $\mathbb{L}$, where objects are different in shape, color, and size. 
The action space of each object is \textit{identical}: \texttt{(north, south, east, west)}.
(2) \textbf{Rush Hour}. To verify the importance of \textit{action binding}, we set each object to have object-specific action space. This variant is motivated by the game \textit{Rush Hour}, where each object is a car and has a (fixed) orientation (only using \textit{triangles}). It can move relatively to its orientation: \texttt{(forward, backward, left, right)}. For example, if a car faces \texttt{east}, \texttt{right} will move \texttt{south}.
For all variants, the action space is factorized by object and has \textit{fixed} order across all scenes, as further discussed in Appendix \ref{sec:action_space}.
More environmental setup is in Appendix~\ref{sec:additional-env}. %

\paragraph{Methods.}
We study the compositional generalization (CG) performance of $6$ approaches, under $3$ categories:
\begin{enumerate}[leftmargin=*]
\item Exact CG: $\symn$-equivariant methods with correct action binding should achieve perfect CG. One such method is \textbf{$\mathbf{\symn}$-CSWM}, the model from~\citet{kipf2019contrastive} with $N$ object masks, one for each library object. Object masks and actions are bound by having the same ordering.
\item No guaranteed CG: To demonstrate the necessity of the three stages in Section~\ref{sec:worldmodel}, we consider methods that either do not have action binding or a $\symn$-equivariant transition model.
In \textbf{C-WM(N)}, we break $\symn$-equivariance by replacing the $N$-slot GNN and shared encoder with a flat MLP.
\textbf{MONet(N)+BM} builds on the model from~\citet{burgess2019monet}, which only extracts objects into masks (slots); we use bipartite matching to align slots $s_t$ and $s_{t+1}$, but \emph{not} to actions, i.e., there is no action binding.
We also include two variants of CSWM that uses only $K$ slots (instead of $N$), but without an explicit binding mechanism: \textbf{$\mathbf{\symk}$-CSWM} assigns each of the $K$ relevant action factors to slots (but the action may not actually control the object in the slot), whereas \textbf{$\mathbf{\symk}$-CSWM(CA)} makes all action factors available to all slots.
\item \edit{Soft} CG: Our method, \textbf{HOWM}, achieves \edit{soft} CG using a $K$-slot approach by relying on the learned action binding (see Section~\ref{sec:worldmodel}).
\end{enumerate}
Additional details about the methods are in Appendix \ref{sec:additional-baselines}.

\paragraph{Training and evaluation setup.}

We follow the setup in~\citet{kipf2019contrastive}, using $1$K episodes for training (consisting of $10$ episodes of length $100$, for $100$ different scenes), and $10$K episodes of length $10$ for evaluation. Additionally, to evaluate compositional generalization, we ensure that (1) the combinations of objects in training dataset are different from those of evaluation dataset, and (2) the training data contains all $N$ objects in the library. Additional training details for each method are in Appendix~\ref{subsec:training}.

Similar to~\citet{kipf2019contrastive}, we measure the dynamics prediction error using two ranking metrics: Hits at Rank 1 (H@1) and Mean Reciprocal Rank (MRR) \cite{kipf2019contrastive}, averaged over $3$ runs. For space, we only report MRR here; H@1 results are similar and can be found in Appendix~\ref{sec:additional_results}. We also report the difference between training and evaluation performance as an indication of the \textit{generalization gap}; this is a proxy for equivariance error, which is difficult to compute in practice due to lack of ground-truth object binding information, as explained in Appendix~\ref{sec:measuring-cg}.

\subsection{Results and analysis}

We compare all methods on the Basic Shapes environment in terms of their generalization performance and scalability, shown in Table~\ref{tab:shapes}.
With sufficient resources, $\symn$-CSWM should be the \textit{upper bound} of performance, since it can achieve perfect CG, as verified by our results (near-perfect eval MRR, near-zero gap).
Both $\symk$-CSWM and $\symk$-CSWM(CA) also achieve excellent training performance since they can memorize the correct action binding in each scene during training;
however, when presented with new scenes in evaluation, the actions may not be bound correctly, leading to worse eval MRR and a large gap.
C-WM(N) lacks $\symn$-equivariance and is significantly worse during both training and evaluation, even with more parameters.
MONet(N)+BM performs even worse, once again indicating the importance of action binding.
Interestingly, it seems to perform better with larger $N$, but also uses much more resources due to its $\symn$-equivariant model;
like $\symn$-CSWM, it exceeds our memory budget for $N = 30$.

HOWM strikes a reasonable middle ground -- the training performance (train MRR) is near-perfect, and generalization performance (eval MRR) is still quite high, though clearly not perfect.
However, in contrast to all other methods except the $\symn$-CSWM upper bound, generalization is maintained as $N$ increases, and the gap is significantly smaller.
These results demonstrate that our proposed approach is able to achieve good generalization for intermediate $N \leq 20$ while consuming significantly fewer resources, and can scale to $N \geq 30$, unlike the exact $\symn$-CSWM method.

One limitation we observed is that for long-term prediction ($\geq 5$ steps), it is still quite challenging to achieve action binding and compositional generalization for a large library of objects. One reason is that the learned representation module (Slot Attention in our case~\citep{locatello2020object}) does not guarantee perfect object representations across time. This affects both action binding learning (if extracted slots are inaccurate, actions cannot be bound correctly) and long-term evaluation (if objects are missed at some step, predictions for all following steps are likely wrong).

\begin{table}[t]
\centering
\caption{
\minor{Results on Rush Hour with relative action space, for $N=5, 10, 20$ (three numbers in each cell).}
}
\vspace{-5pt}
\label{tab:rushhour}
\scriptsize
\begin{tabular}{l|ll|l}\toprule

\textbf{Env=Rush Hour}  & \textbf{MRR} (1-step) & \textbf{MRR} (5-step)  & \textbf{Gap}  (MRR, 5-step)  \\
\midrule
$\Sigma_N\text{-CSWM}$ & 100.\; 100.\; 99.9 & 99.9\; 99.8\; 99.8 & \; 0.1\; \; 0.2\; \; 0.1 \\
\midrule
$\Sigma_K\text{-CSWM}$ & 100.\; 66.9\; 47.6 & 99.9\; 29.5\; 13.7 & \; 0.1\; 66.0\; 85.0 \\
$\Sigma_K\text{-CSWM(CA)}$ & 99.2\; 55.4\; 80.2 & 94.9\; 15.5\; 15.8 & \; 4.2\; 84.5\; 84.2 \\
$\text{C-WM(N)}$  & 55.5\; 67.8\; 80.6 & 23.7\; 24.5\; 45.3 & 48.8\; 70.7\; 54.1 \\
\midrule
$\textbf{HOWM}$ \textbf{(ours)}  & 96.7\; 94.3\; 98.7 & 84.3\; 63.2\; 65.3 & 11.2\; 31.1\; 31.3 \\
\bottomrule
\end{tabular}
\vspace{-15pt}
\end{table}

In the Rush Hour environment, shown in Table~\ref{tab:rushhour}, the action-binding problem is more challenging and requires more sophisticated modeling of object-dependent transition dynamics. We report the evaluation MRR and the generalization gap. As before, $\symn$-CSWM acts as an upper bound and performs near-perfectly with near-zero gap; however, as before, we expect that it will not scale beyond $N=20$ with our resource budget.
All other methods, including HOWM, have worse performance compared to Shapes. However, HOWM still generalizes significantly better to new scenes for $N=10$ and $N=20$, whereas performance of other methods declines more steeply.

\subsection{\edit{Visualization}}

\begin{figure}[t]
\centering
\includegraphics[width=0.43\textwidth]{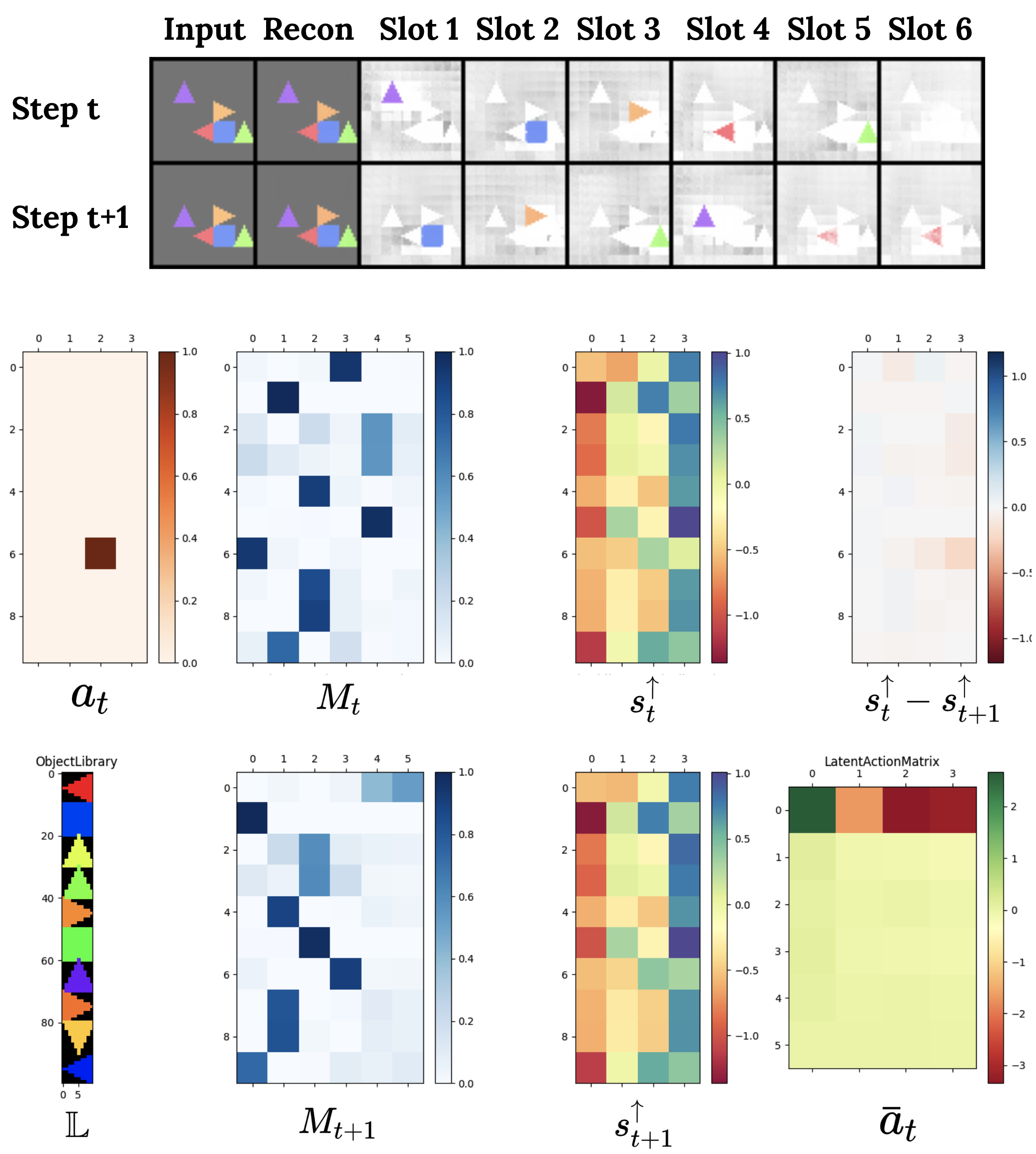}
\vspace{-0.1in}
\caption{
\
\edit{
\textbf{(Top)} Visualization of learned slots; note that they are in random ordering.
\textbf{(Bottom)} Components from a learned model for a transition, with $N=10$ (corresponds to rows besides $\bar{a}_t$) and $K=5$ (corresponding to $6$ columns with an additional slot for background).
}
}
\label{fig:vis-combined}
\vspace{-0.3in}
\end{figure}

\edit{
We visualize a model on a sampled transition $(s_t, a_t, s_{t+1})$ with $N=10$ and $K=5$ in Figure \ref{fig:vis-combined}.
Action attention matrices are computed using the states: $M_t(s_t),M_{t+1}(s_{t+1}) \in \mathbb{R}^{N \times (K+1)}$.
The visualized $M$ align with expectation: (1) $K$ present objects are bound to $K$ slots.
(2) the absent $N-K$ objects are randomly assigned to slots (reasonable since not forced by loss), but they are consistent in $t$ and $t+1$ (as forced by loss to keep $N$-object $s_t^\uparrow, s_{t+1}^\uparrow$ embeddings close).
Also, the lifted embeddings $M_t^+ \bar{s}_t,M_{t+1}^+\bar{s}_{t+1}$ are well aligned and have little difference.
}

\section{Related work}

\textbf{Object-oriented representations} are crucial in artificial intelligence and robotics \citep{diuk2008object,DBLP:phd/ndltd/Wong16}; our framework is related to OO-MDPs~\citep{diuk2008object} consisting of a single class.
Recently, a line of works studies learning to \textit{discover} objects and their representations end-to-end.
MONet \citep{burgess2019monet} applies sequential attention with VAEs to learn factorized representations.
GSWM \citep{lin2020improving} builds generative models and learns end-to-end with variational inference.
Slot Attention \citep{locatello2020object} proposes attention at the pixel level, which can be viewed as soft clustering of pixels of objects.
The order of object slots is decided by randomly initialized cluster centroids.

A object-oriented world model can be learned jointly or separately with object representation, using pixel reconstruction (COBRA \citep{watters2019cobra}, OAT \citep{creswell2021unsupervised}), variational inference (STOVE \citep{kossen2019structured}, OP3 \citep{veerapaneni2020entity}), or contrastive loss (C-SWM \citep{kipf2019contrastive}, NPS \citep{didolkar2021neural}).
Furthermore, in jointly learning object representations and world models, a key problem is \textit{object binding} \citep{greff2020binding}: \edit{binding object \textit{slots} between time steps}, which is typically solved by bipartite matching.
In our setup, we emphasize the importance of correctly \textit{binding actions} to objects.
\edit{In video prediction such as \citep{kipf2021conditional}, action is not considered and thus has no action binding issue, which is our key focus and brings the major challenge.}
\edit{
\citet{biza2022binding} use location-based action space and learn attention to factorize monolithic actions. It keeps the setup in CSWM with $N=K$ and does not do compositional generalization in our $N>K$ formalism.
}
\minor{We further discuss the choice of action space in Appendix \ref{sec:action_space}.}

\textbf{Compositional generalization} has been widely studied in deep language models \citep{johnson2017clevr}.
It also known as \textit{systematic generalization} \citep{bahdanau2018systematic}, and \textit{systematicity} \citep{lake2018generalization}.
Similar to us, \citet{gordon2019permutation} also use permutation equivariance to model compositional generalization in supervised learning on language data.
\citet{keysers2019measuring} propose principles of measuring compositional generalization on sentences.
A similar concept, \textit{compositionality}, is usually referred to in disentangled representation learning. However, it typically focuses on disentanglement between individual attributes of distributed representations \citep{higgins2018towards,caselles2019symmetry}, while we are more interested in factorization between (attributes of) objects.
There is also work on measuring compositionality \citep{andreas2019measuring,chaabouni2020compositionality}.
\edit{Beyond language}, \citet{locatello2020object,veerapaneni2020entity} learn object-oriented representations, whose compositional generalization usually refers to factorization by objects.
Furthermore, no prior work considers the difference between present objects $K$ and its ``library'' $N$ objects, i.e., they can only use $\symn$-equivariant model and thus struggle in scaling up.

\textbf{Symmetries and MDP homomorphisms.}
Symmetry widely exists in various real-world data \citep{bronstein2021geometric}, and also in MDPs \citep{ravindran2004algebraic}.
A line of works in equivariant networks studies equivariance properties of existing network architectures, such as permutation equivariance in graph neural networks (GNNs)~\citep{keriven2019universal}, while some other works study equivariant constraints to other symmetry groups \citep{bronstein2021geometric}.
In RL, symmetries in MDPs can be modeled by MDP homomorphisms \citep{ravindran2004algebraic}, which have nice properties such as \textit{optimal value equivalence} and can be exploited in policy learning~\citep{van2020mdp}.

\section{Conclusion}

In conclusion, in this paper we introduced the Object Library family of object-oriented environments, as a first step toward studying compositional generalization in reinforcement learning. We defined compositional generalization in Object Library using the language of permutation equivariance, and showed a construction involving a $K$-slot MDP that was shown to achieve compositional generalization in Object Library. Based on this analysis, we introduced a three-stage pipeline for learning an object-oriented world model, and demonstrated in two instances of Object Library environments it could indeed generalize well to libraries with large $N$, while consuming fewer resources. 
Nevertheless, there remains a \edit{performance} gap between our models and true compositional generalization that requires further study.
Additionally, our analysis has been based on a fixed $K$, and assumes object persistence \edit{and a single object class}, which clearly should be relaxed.
Finally, while we have focused on learning transition models with compositional generalization, their application to planning and sequential decision-making should be explored.

\section*{Acknowledgments}

This work was supported by NSF Grants \#2107256 and \#2134178. 
R. Walters is supported by The Roux Institute and the Harold Alfond Foundation. We also thank Ondrej Biza for helpful discussions and anonymous reviewers for useful feedback.

\bibliography{reference}
\bibliographystyle{icml2022}

\newpage
\appendix
\onecolumn
\section*{Appendix}

\section{Outline}

We discuss potential extensions and the choose of factorized action space in Section \ref{sec:discussion}.
We provide some additional details of our framework for compositional generalization in Section \ref{sec:additional_framework}, followed by the proof of our main theoretical result, Proposition \ref{prop:error}, in Section \ref{sec:proofs}.
We further discuss the measure of compositional generalization in Section \ref{sec:measuring-cg} and HOWM in \ref{sec:howm_details}.
Additional results, along with details about experiments and environments, are in Section \ref{sec:additional_results}, Section \ref{sec:additional-env}, and Section \ref{sec:appendix-experiment}, respectively.

\section{Additional Discussion}
\label{sec:discussion}

\subsection{Potential extensions}
\label{sec:extension}

Our framework makes some assumptions in modeling the Object Library.
For example, we assume object persistence, i.e., we have $K$ specific objects in every scene.
We can relieve this assumption, and thus needs to consider compositional generalization to different $K$'s and objects changing in each scene.
However, these assumptions are not fundamental limitations of the framework, and we provide some straightforward ideas to extend the framework.

\paragraph{Extension: reward and planning.}
Our framework can also handle reward. We do not include that, since we do not focus on using the learned model (in the slot MDP $\slotmdp$) for planning.
Additionally, in other words, the model can also learn task-relevant features or objects that preserves the structure of reward and transition.

\paragraph{Extension: generalization to different $K$.}
Our framework can extend to compositional generalization to \textit{different} $K$'s.
For a fixed $K$ in the paper, there is a unique slot MDP $\slotmdp$ that is factored MDP and has permutation automorphism on it.
However, for different $K$ (usually training on smaller $K$'s and generalizing to larger $K$'s), there should be multiple slot MDPs for each K.
Therefore, there would be an extra step to build a structured homomorphism between different slot MDPs first, $\mathcal{M}_{[K_i]}$ and $\mathcal{M}_{[K_j]}$.
The model can then generalize to different $K$'s by first generalizing to a slot MDP of the corresponding $K_j$ (by recomposing structured homomorphism).

\paragraph{Extension: changing objects.}
The framework can also be extended to handle changing of objects during manipulation.
We can split the episode to two episodes at the point of object changing.
If the number of objects does not change, it is simply another scene.
Otherwise, generalization to different values of $K$ may need to be considered.

\paragraph{Extension: object class.}
In our framework, we implicitly assume that we just have one type of objects, or \textit{class}, i.e. all interaction between objects can be modeled by one type of edge layer in graph neural networks.
The concept of class is introduced in OO-MDPs~\citep{diuk2008object}; our framework can be considered a single-class OO-MDP with multiple object instances of the same class. In contrast to OO-MDPs, our states have continuous attributes (instead of Boolean ones).
Additionally, we perform representation learning from the image-based MDP $\pixelmdp$, thus the learned object representation can be imperfect and more importantly, each individual slot can represent all objects and thus can represent different objects across time steps.

\subsection{Factorized vs. location-based action space}
\label{sec:action_space}
In our work, we assume the action space is factorized by objects, similar to the state space.
Empirically, this can be viewed as a high-level abstraction of the actions, such that we can independently control each object.
However, there is another option: \textit{location-based action space}, which chooses the object to control by location, like a touchscreen.
We argue that it is inherently not a good choice for our setup.
In the implementation, we require object-based world modeling to have (1) temporal consistency, and (2) compositional generalization. 
For location-based action space, these two requirements \textit{cannot} be satisfied at the same time, even though location-based actions implicitly provide transition and thus have advantages.

\textit{Option (1) universal slots}: temporal consistency issue.
One option is that the extractor can bind any objects to any slots. However, this results in random slot ordering even between two adjacent time steps.
Thus, the model needs to solve the correspondence between slots in two time steps. But because of location-based action space, there is no canonical ordering of either $K$ present objects or $K$ slots.
In other words, the transition modeling has to rely on iterative approaches, such as bipartite matching, to optimize for a potentially correct order for every step, which is computationally expensive and does not necessarily give correct answer, especially for long-term prediction.

\textit{Option (2) dedicated slots}: compositional generalization issue.
Another option is to fix the ordering in object extractor, so for every scene the extractor outputs object slots in \textit{fixed} order. This is the case of our $N$-slot C-SWM variant.
Unfortunately, this can only achieve perfect results in $N=K$ as the original paper does \citep{kipf2019contrastive}, while it does not work for general case $N>K$, since there must be a slot that needs to bind to more than one object.
However, we do \textit{not} have a canonical ordering of all $N$ objects in the library, so it is impossible to guarantee that for every seen or unseen scenes, it is able to provide dedicated and consistent slots across all time steps and all scenes.

If we do not consider computational efficiency, we can also feed actions of all objects to the transition model, i.e., no factorization of action.
However, it becomes less meaningful for factorizing into $N$ object slots and the model needs $O(N^2)$ complexity to model $N$ object slots and $N$ actions for every object slot.

Overall, factorized action space has several technical advantages and can be intuitively understood as a set of skills that independently control each object, which has been widely used in factored MDPs \cite{ravindran2004algebraic,guestrin2003efficient,sutton2018reinforcement} and robotics applications \cite{kroemer2019review,kipf2019contrastive,veerapaneni2020entity,biza2021impact}.

\section{Measuring Compositional Generalization in Object Library}
\label{sec:measuring-cg}

By compositional generalization, we aim to generalize to novel object compositions with seen objects.
This means that we train a model in slot MDP $\slotmdp$ using data from some scenes, and want to measure the performance on other scenes.
In this process, one main error comes from the binding side $\slotmdp \mapsto \scenemdp$: (1) from training scenes to the slot MDP, and (2) from the slot MDP to isomorphic novel scenes.
Therefore, to estimate the performance, we need to have two set of scenes that are different compositionally, to fully measure the error.

To this end, we propose a strategy to generate training and evaluation data for dynamics prediction, which also appears similarly in generating compositional natural language data \cite{keysers2019measuring}.
(1) \textit{Disjoint scene object set: $\{i_k: i_k \in [N], k \in [K]\}=\sceneset_j \in \mathcal{P}_K(\libset)$.} Scenes should sampled uniformly and different in training and evaluation set (not a trivial permutation).
(2) \textit{Similar object distribution: $i_n \in \libset$.} Both datasets should have similar objects, to ensure the error is not biased by errors in object detection.
In the experiment section, we use this strategy to collect data and measure the performance between $\hat{T}_\slots$ and $T_\slots$ directly in latent space.

\paragraph{Difficulty of numerically computing equivariance error.}

We use the above strategy to measure generalization performance in predicting action effects in novel scenes. Our theoretical framework technically provides another metric $\lambda$, Definition~\ref{def:equivariance-error} in the main text, to measure compositional generalization.
Although this metric is useful for proving our main theoretical result, it typically provides vacuous results for deterministic domains (e.g., the environments we use in our experiments), because the predicted transition function will always have an equivariance error of $\lambda = 1$, unless the transition function is exactly correct, only in which case $\lambda = 0$. Unfortunately, due to small prediction errors and numerical inaccuracies, empirically we always get $\lambda = 1$.

One alternative for deterministic domains is to instead compute the error in the (latent) \emph{state} space (i.e., compare $\phi(s')$ vs. $\phi(\sigma.s')$), instead of in the transition probabilities. However, this raises another issue related to \emph{binding}. In our slot-based framework, the ordering of slots is arbitrary, so the same dimensions of $\phi(s')$ and $\phi(\sigma.s')$ may not be comparable. Accurate comparison requires solving the binding problem, hence a metric computed in this fashion is approximate at best. Instead, we resort to simply measuring prediction performance using the strategy described above.

\minor{To the end, we chose to use \textit{compositional generalization gap}, reported as a metric in the tables.}

\section{Additional Description of Homomorphic Object-oriented World Models}
\label{sec:howm_details}

\begin{figure*}[t]
\centering
\subfigure{
    \includegraphics[width=0.59\linewidth]{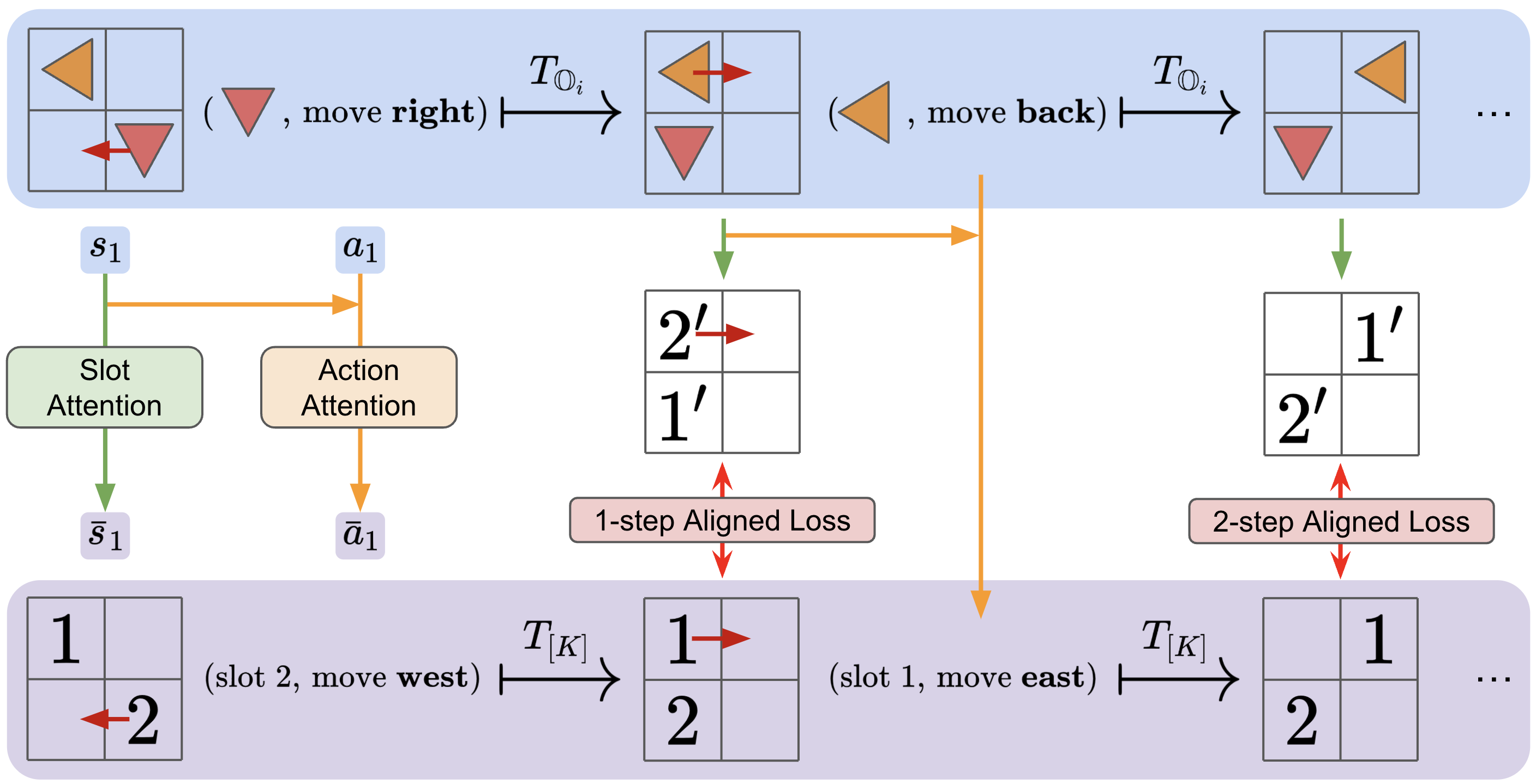}
}
\hfill
\subfigure{
    \includegraphics[width=0.31\linewidth]{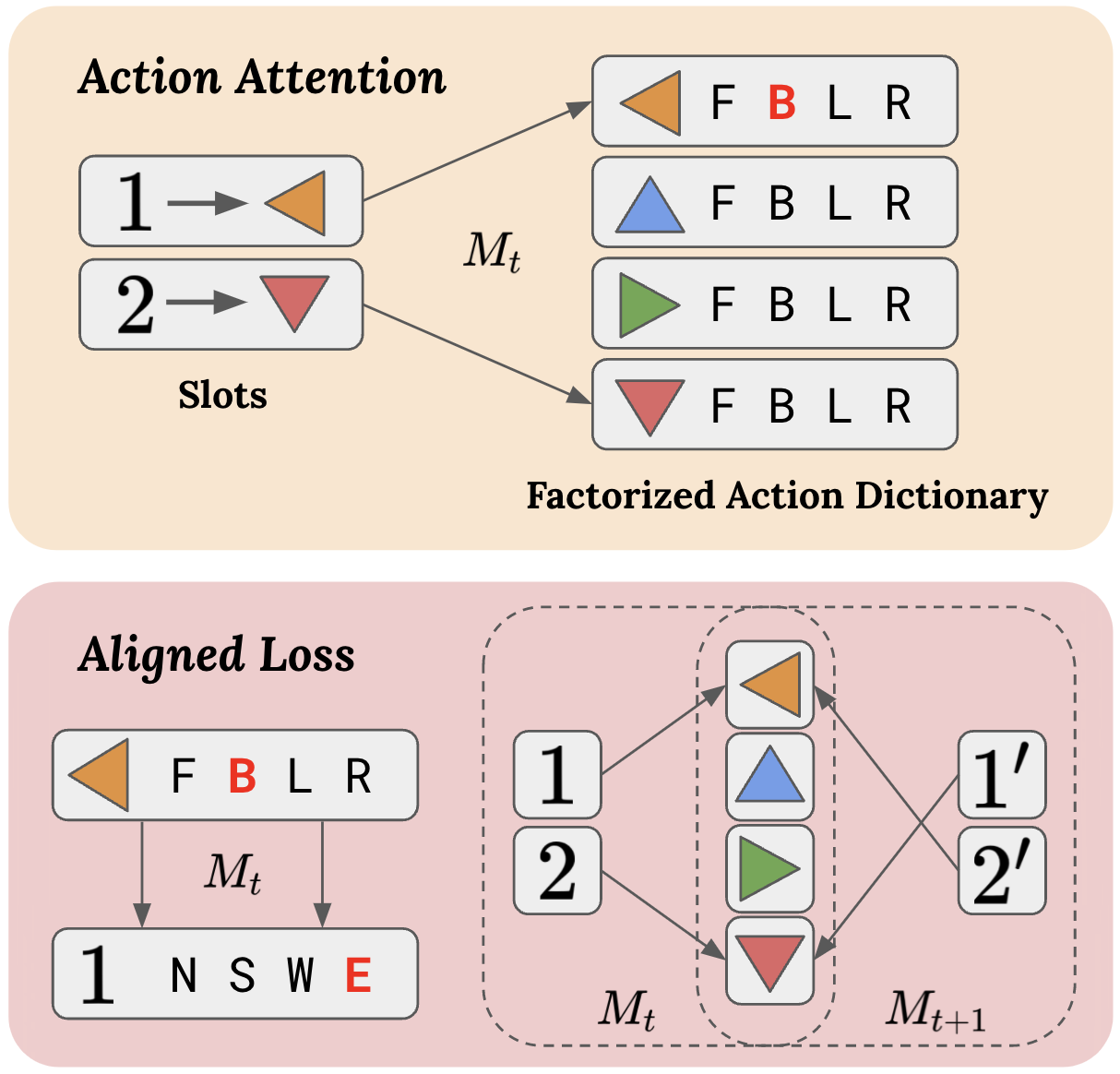}
} 
\centering
\caption{
A copy of the figure in main text.
}
\vspace{-10pt}
\label{fig:method_overview_copy}
\end{figure*}

We extend the idea behind two proposed components: \textbf{Action Attention} and \textbf{Aligned Loss} to explain how the latent world model is able to train end-to-end, and also bring more details about \textbf{multi-step model prediction} in latent space which is not further elaborated in the main text.

The key challenge comes from the object-orientation representation side, where no canonical or unique order of objects can be defined in images.
Thus, we have two potential paths to deal with this.
One path is to use information from actions and the interaction with environment $(s_t,a_t,s_{t+1})$, since the action is usually independently controlling one or a few entities in the environment.
Another path can be simply aligning the slots (in different order) across different time steps $(s_t, s_{t+1}, s_{t+2}, s_{t+3},\ldots)$, which is the idea behind using bipartite matching or Hungarian algorithm and has been used in recent works for differentiable bipartite matching, for object detection and video understanding.

\subsection{Contrastive Aligned Loss}
We are based on the object-structured contrastive loss as described in \citep{kipf2019contrastive}.

Since we have $K$ slots and $N$ possible objects in the library, a slot has to bind to more than one object. Thus, the order of the slots must be different for scenes with $K$ objects from the library.
Then, the problem is that, we cannot compute the loss directly between two slots in adjacent time step (in one action step away).
To solve this without invoking bipartite matching between adjacent states $(s_t, a_t, s_{t+1})$, we need to make use of some canonical order. We assume there exists a fixed (across all scenes, guaranteed at data generation time in implementation) but unknown order of $N$ objects in the library. Thus, there is an imaginary canonical full MDP, where the order of the factorized state and action space is fixed (but unknown and arbitrary).

The intuition to make use of this is straightforward:
(1) in predicting the next state, we compute in the slot MDP $T(\bar{s}_t, \bar{a}_t)$ using $\Sigma_K$-equivariant GNN to lower the cost.
(2) in computing the loss, we map the predicted latent states and target latent states (both represented in $K$ slots but potentially in different order) back to the canonical full MDP, to ensure the temporally consistent order between adjacent steps $(s_t, a_t, s_{t+1})$. Intuitively, other unused $N-K$ "slots" (factorized state spaces) can be zeros.

In summary, this allows us to jointly learn (1) factorized (object-oriented) representations for $N$ objects and (2) a latent $\Sigma_K$-equivariant transition model, by utilizing a factorized action space with fixed, arbitrary, unknown order.

Similar to \citep{locatello2020object}, the softmax is over $K$ slots, where the intuition is that every object should select at most one slot. Thus, we also include an additional slot, for non-present objects and also for encoding background.

\begin{equation}
M_t (s_t) = \underset{K}{\operatorname{softmax}}\left(\frac{1}{\sqrt{D}} k\left( \mathrm{Id}_{N \times N} \right) \cdot q\left( s_t \right)^{T}\right) \;,
\end{equation}
where $k, q$ are linear projections mapping to some dimension $D$, and $a_t$ acts as the value $v$ without further transformation.
Also, we only allow the gradient from Action Attention.

\subsection{Multi-step Evaluation}

We omit the details of multi-step evaluation in the main text, while this is the important component for how the model works for long-term prediction.
The key idea behind multi-step evaluation is generic, where we learn a MDP homomorphism $h: \mathcal{M} \to \overline{\mathcal{M}}$, and unroll in the reduced MDP $\overline{\mathcal{M}}$ and somehow lift the prediction using learned model in $\overline{\mathcal{M}}$ back to the ground MDP ${\mathcal{M}}$.
The main difference is that we only need the information of ordering, where the project property guarantees that it is somehow preserved, thus the lifting is possible (only keeping the ordering correct, by aligning object slots to some canonical object order).

The key idea is to achieve prediction in latent space (slot MDP) $\bar{T}$, where for simplicity we use $T$ below.
We introduce how we achieve that.
When we need to sequentially query the transition model to give us rollouts of the future in some latent slot space (i.e., the $K$-slot MDP), a problem comes: we map every image in a trajectory to a sequence of slots, while there is no mechanism enforcing temporal consistency, i.e., different slots in different time step can bind to different objects.
So, we cannot use $T(...T(T(s_1,a_1), a_2),...)$ to iteratively rollout the model directly.
We propose a strategy that can rollout the model in latent space (the $K$-slot MDP) directly, without using bipartite matching or accumulated error of sequentially matching adjacent steps, under some assumptions:
(1) we know that the factorized action space has a fixed but \textit{unknown} order, and (2) the project property holds, i.e. the binding exists and object identities are preserved (we can identity how objects replacement corresponds to the permutation of slots).
    
The reason we can't align states directly is that, we need sequentially unrolling the transition model, which will result in using the binding matrix at each time step and easily have accumulated error (by multiple a sequence of learned approximate binding matrices).

Let's say $M_{t} M_{t+k}^+ = P_k \in \mathbb{R}^{K\times K}$, denoting the alignment to $k$ steps ahead using learned binding $M_t$, which maps the slots to the canonical space (full MDP) and back to a specific K-slot ordering.
If the project property satisfies, it should be full rank.
Thus, for predicting next step, it is $T(\bar{s}_t, \bar{a}_t) \approx P_1\bar{s}_{t+1}$, and similarly for $T(\bar{s}_{t+1}, \bar{a}_{t+1}) \approx P_{1\to 2}\bar{s}_{t+2}$.

Because of the $\Sigma_K$-equivariance property of the transition network, the prediction $M_t^+ T(\bar{s}_t, \bar{a}_t) \approx M_{t+1}^+\bar{s}_{t+1}$ is equivalent to transforming actions: $M_{t+1} M_t^+ (T(\bar{s}_t, \bar{a}_t))\approx \bar{s}_{t+1}$ and $T(\bar{s}_t, \bar{a}_t)\approx M_{t} M_{t+1}^+ \bar{s}_{t+1}$.

Similarly, for next step, we would need to align for the next step again:
\begin{align}
 T( M_{t} M_{t+1}^+ \bar{s}_{t+1}, M_{t} M_{t+1}^+ \bar{a}_{t+1} ) & =  (M_{t} M_{t+1}^+) (M_{t+1} M_{t+2}^+) \bar{s}_{t+2} \\
   \iff  T( M_{t} M_{t+1}^+ \bar{s}_{t+1}, M_{t} M_{t+1}^+ \bar{a}_{t+1} ) & = ( M_{t} M_{t+2}^+ ) \bar{s}_{t+2} \\
   \iff T( T(\bar{s}_t, \bar{a}_t), M_{t} M_{t+1}^+ \bar{a}_{t+1} ) & = M_{t} M_{t+2}^+ \bar{s}_{t+2} \\
   \iff T( T(\bar{s}_t, \bar{a}_t), P_1 \bar{a}_{t+1} ) & = P_2 \bar{s}_{t+2}
\end{align}

Telescoping to $k$ steps ahead, we further get
\begin{align}
T(\ldots(T(\bar{s}_t, \bar{a}_t), (M_{t} M_{t+1}^+) \bar{a}_{t+1} )\ldots,(M_{t} M_{t+k}^+) \bar{a}_{t+k}) &= (M_{t} M_{t+k}^+) \bar{s}_{t+k} \\
\iff T(\ldots(T(\bar{s}_t, \bar{a}_t), P_1 \bar{a}_{t+1} )\ldots,P_k \bar{a}_{t+k}) &= P_k \bar{s}_{t+k}.
\end{align}

Finally, we only need to map it to the $N$-object canonical full MDP:
\begin{align}
    M_{t} (T(\ldots(T(\bar{s}_t, \bar{a}_t), (M_{t}^+ M_{t+1}) \bar{a}_{t+1} )\ldots,(M_{t}^+ M_{t+k}) \bar{a}_{t+k}) ) &= M_{t+k} \bar{s}_{t+k} \\
    \iff M_{t} (T(\ldots(T(\bar{s}_t, \bar{a}_t), P_1 \bar{a}_{t+1} )\ldots, P_k \bar{a}_{t+k}) ) &= M_{t+k} \bar{s}_{t+k}.
\end{align}

However, this requires us to align sequentially with $P_1, P_2, \ldots, P_k$ for k-step prediction, which could not scale for large $k$.
In actual implementation, we instead \textit{align actions} to the order of $s_t$.
Recall that the latent action is computed by $\bar{a}_t = M_t(s_t) a_t$, thus we have 
\begin{equation}
    (M_{t} M_{t+k}^+) \bar{a}_{t+k}  = P_k \bar{a}_{t+k} = M_{t} a_{t+k}. 
\end{equation}

In computing the evaluation metric for multi-step prediction, such as $k$-step Hits and MRR, we could compute as follows:
\begin{equation}
    M_t^+ \left(T(\bar{s}_t, [M_t a_t, M_{t} a_{t+1}, ..., M_{t} a_{t+k-1}]) \right) \approx M_{t+k}^+ \bar{s}_{t+k},
\end{equation}
where $[M_t a_t, M_{t} a_{t+1}, ..., M_{t} a_{t+k-1}]$ means that we sequentially input $k-1$ actions to the world models and repressively predict next state using last state for $k$ times.
Intuitively, we do not need to know intermediate binding $M_{t+1}$ through $M_{t+k-1}$ since the actions are in a canonical order (defined by the N-object full MDP) and we align all intermediate slots to the slot order of first state $s_t$.

We use slot size of $16$ and embedding size of $4$ for all reported numbers.
The negative states are sampled half from scenes with same objects $\mathbb{O}_i$, and half from different scenes, where we empirically find it works better than using purely random episodes, as further explained in Section \ref{sec:additional_results}.

\section{Details of the Compositional Generalization Framework}
\label{sec:additional_framework}

Symmetry widely exists in various real-world data \citep{bronstein2021geometric}, and also in MDPs \citep{ravindran2004algebraic}.
A line of works in equivariant networks studies equivariance properties of existing network architectures, such as permutation equivariance in graph neural networks (GNNs)~\citep{keriven2019universal}, while some other works study equivariant constraints to other symmetry groups \citep{bronstein2021geometric}.
The framework of \textit{MDP homomorphism} is an algebraic approach studying \textit{symmetries} and \textit{abstraction} in reinforcement learning \citep{ravindran2004algebraic}.
MDP homomorphisms can be \textit{induced} by symmetric stucture in MDPs \citep{ravindran2004algebraic,van2020mdp}, such as spatial transformations like reflections.
We use the framework to study object replacement symmetry for formulating compositional generalization.
Homomorphic MDPs have other nice properties to be further explored, such as \textit{optimal value equivalence} \citep{ravindran2004algebraic}.
Also, the lifted policy learned in homomorphic MDPs has equivariance properties, which can be exploited by equivariant networks \citep{van2020mdp}.

\subsection{Definition with Reward}
\label{subsec:reward}
In the main paper, we focus on the equivariance of \textit{transition} function.
We provide the full definition of \textit{error} also with \textit{reward} function, another condition in MDP homomorphisms.
The error of reward function only needs \textit{invariance}, since it is a mapping with real-valued output: $R: \mathcal{S} \times \mathcal{A} \to \mathbb{R}$.

\begin{definition}[Invariance error of $\hat{R}_\libset$ on $\fullmdp$]
Let $\hat{R}_\libset: \mathcal{S}_\libset \times \mathcal{A}_\libset \to \mathbb{R}$ be a (learned) reward model of $\fullmdp$.
The invariance error $\lambda _ {R}$ of $\hat{R}_\libset$ with respect to operation $p_\libset$ is defined as:
\begin{equation}
\lambda  _ {R} \triangleq
\mathop{\mathbb{E}}\limits_{\sigma \in \symn, (s,a) \in \mathcal{S}_\libset \times \mathcal{A}_\libset}
\left[
\left| 
\hat{R}_\libset(s,a) - \hat{R}_\libset( \sigma.s, \sigma.a) 
\right|
\right] \;,
\end{equation}
where $\symn$ is the permutation group of size $N$, and $\libset$ is the object library of size $|\libset|=N$.
\end{definition}

The definition of the error on reward only needs invariance.
The extended version of Proposition \ref{prop:error} including reward and its proof directly follow the one for transition function.

\subsection{Definition of Induced Operations}
\label{subsec:induced_operations}

We formally define the induced operation on scene and illustrate it with diagrams.

\begin{definition}[Object replacement operation $p_\libset$]
\label{def:obj_rep_op}
Let $\symn$ denote the permutations of $N$ elements, which acts naturally on the object library set $\libset$.
The object replacement operation is the group operation $p_\libset: \symn \times \libset \to \libset$ where $\symn$ acting on $\libset$ to replace the identity of objects.
\end{definition}
For example, using disjoint cycle notation, an object replacement operation $\pi = (1 2 3)(45)$ acts on object indices in $\libset$ by sending $\pi(3)=1$ or $\pi(4) = 5$.
Intuitively, this operation permutes the label or index of objects in $\libset$, thus the scene has identical appearance but with object labels being replaced.
This induces a group operation on scenes, which can be defined below.

\begin{definition}[Induced scene set operation $p_K$]
\label{def:scene_op}
The permutation $\symn$ on object library $\libset$ \textit{induces} an operation transforming scene object sets $\sceneset \subset \libset$.
The induced group operation $p_K: \symn \times \mathcal{P}_K(\libset) \to \mathcal{P}_K(\libset)$ acts on the set $\mathcal{P}_K(\libset)$ as such $p_K(\pi,\{ i_1, \ldots, i_K \}) = \{ \pi(i_1), \ldots, \pi(i_K) \}$, where $\pi \in \symn$, object indices $\{ i_1, \ldots, i_K \}, \{ \pi(i_1), \ldots, \pi(i_K) \} \in \mathcal{P}_K(\libset)$, each object index $i_k \in [N], K < N$.
\end{definition}
Since $i_k$ is the index of objects, $p_K$ maps a set of objects to another set indexed by $\{ \pi(i_1), \ldots, \pi(i_K) \}$.
If $\pi=(123)(45), K=2$, then $p_K(\pi,\{ 3,4 \}) = \{ 1,5 \}$. 
The action of $\Sigma_N$ on $\mathcal{M}_\mathbb{L}$ defined in Section \ref{subsec:ooe} is compatible with the action of $\Sigma_N$ on $\mathcal{P}_K(\libset)$ as $\sigma o(s) = o (\sigma s)$.  Thus, $\Sigma_N$ permutes the scene MDPs $\lbrace \mathcal{M}_\sceneset \rbrace$.
The induced operation on scenes implies some symmetry between sub-MDPs of scenes $\mathcal{M}_\sceneset$ consisting of objects $\sceneset$, which can be naturally studied by \textit{MDP homomorphisms} between scene MDPs $\{ \scenemdp: \sceneset \in \mathcal{P}_K(\libset) \}$.

\subsection{Illustrative Examples}

Graphically, the induced operation on scene can be seen as below.

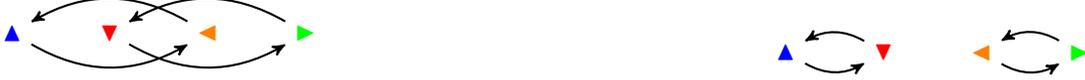
\begin{figure}[t]
\hspace{30pt}
    \subfigure{
        \begin{tikzpicture}[->,>=stealth',auto,node distance=1.3cm,
          thick,main node/.style={circle,draw,font=\sffamily\Large\bfseries}]
        
          \node (1) {\textcolor{blue}{$\blacktriangle$}};
          \node (2) [right of=1] {\textcolor{red}{$\blacktriangledown$}};
          \node (3) [right of=2] {\textcolor{orange}{$\blacktriangleleft$}};
          \node (4) [right of=3] {\textcolor{green}{$\blacktriangleright$}};
        
          \path[every node/.style={font=\sffamily\small}]
            (1) edge [bend right] node  {} (3) 
            (3) edge [bend right] node  {} (1)
            (2) edge [bend right] node  {} (4) 
            (4) edge [bend right] node  {} (2);
        \end{tikzpicture} 
    } 
    \hfill
    \subfigure{
        \begin{tikzpicture}[->,>=stealth',auto,node distance=1.3cm,
      thick,main node/.style={circle,draw,font=\sffamily\Large\bfseries}]
    
          \node (1) {\textcolor{blue}{$\blacktriangle$}};
          \node (2) [right of=1] {\textcolor{red}{$\blacktriangledown$}};
          \node (3) [right of=2] {\textcolor{orange}{$\blacktriangleleft$}};
          \node (4) [right of=3] {\textcolor{green}{$\blacktriangleright$}};
        
          \path[every node/.style={font=\sffamily\small}]
            (1) edge [bend right] node  {} (2) 
            (2) edge [bend right] node  {} (1)
            (3) edge [bend right] node  {} (4) 
            (4) edge [bend right] node  {} (3);
        \end{tikzpicture}
    } 
    \hspace{30pt}
    \caption{
    Examples of permutations on a library $\libset = $  \{\textcolor{blue}{$\blacktriangle$},
\textcolor{red}{$\blacktriangledown$},
\textcolor{orange}{$\blacktriangleleft$},
\textcolor{green}{$\blacktriangleright$}\}.
    }
    \label{fig:fig-permutation}
\end{figure}

Before introducing permutation groups, we first introduce how to use cycle notation to represent permutations. Assume our object library $\libset = $  \{\textcolor{blue}{$\blacktriangle$},
\textcolor{red}{$\blacktriangledown$},
\textcolor{orange}{$\blacktriangleleft$},
\textcolor{green}{$\blacktriangleright$}\}. 
The permutation in Figure \ref{fig:fig-permutation} left is denoted by the cycle notation (\textcolor{blue}{$\blacktriangle$}\textcolor{orange}{$\blacktriangleleft$})(\textcolor{red}{$\blacktriangledown$}\textcolor{green}{$\blacktriangleright$}).      
The permutation in Figure \ref{fig:fig-permutation} right is denoted by the cycle notation (\textcolor{blue}{$\blacktriangle$}\textcolor{red}{$\blacktriangledown$})(\textcolor{orange}{$\blacktriangleleft$}\textcolor{green}{$\blacktriangleright$}). 
A permutation can be seen as an action that rearranges a set of elements. A permutation group of a set $A$ is a set of permutations of $A$ that forms a group under function composition.

We can use the symmetric group $\symn$ to replace any object in the scene $\mathbb{O}_i$. For example, assume $N=4$, $K=2$ and scene $\mathbb{O}_i$ consists of objects \{ {\textcolor{orange}{$\blacktriangleleft$}, \textcolor{red}{$\blacktriangledown$}} \}. 
Then the group operation on $\mathbb{L}$ ((\textcolor{red}{$\blacktriangledown$} \textcolor{green}{$\blacktriangleright$}), \{\textcolor{blue}{$\blacktriangle$},
\textcolor{red}{$\blacktriangledown$},
\textcolor{orange}{$\blacktriangleleft$},
\textcolor{green}{$\blacktriangleright$}\}) $\mapsto$ \{\textcolor{blue}{$\blacktriangle$},
\textcolor{green}{$\blacktriangleright$},
\textcolor{orange}{$\blacktriangleleft$},
\textcolor{red}{$\blacktriangledown$}\} is essentially doing the object replacement: $\sigma$  (\textcolor{red}{$\blacktriangledown$}) = \textcolor{green}{$\blacktriangleright$}. Also, it can be viewed as scene replacement: $R$: \{ {\textcolor{orange}{$\blacktriangleleft$}, \textcolor{red}{$\blacktriangledown$}} \} $\mapsto$  \{ \textcolor{orange}{$\blacktriangleleft$}, \textcolor{green}{$\blacktriangleright$}\} .

\section{Proof of Proposition 4.1: Scaled Equivariance Error}
\label{sec:proofs}

We provide the proof of Proposition \ref{prop:error} in this subsection. 

Although we have expectation in the definition, the relationship holds \textit{point-wisely}, i.e., for every permutation $\sigma \in \symn$, the equivariance error is magnified by some constant $C={N \choose K}$ in the slot MDP.
We denote the error quantity of a permutation $\sigma$ as $\lambda_\slots^{\overline{\sigma}}$ (for slot MDP) and $\lambda^\sigma_\libset$ (for full MDP).

Note that the sample equivariance error is
\begin{equation}
\lambda_\slots^{\overline{\sigma}}
=^{(1)} \left| 
\hat{T}_\slots(\phi\left(s^{\prime}\right) \mid \phi(s), 
\alpha_{s}(a)) 
- 
\hat{T}_\slots(\bar{\sigma}. \phi\left(s^{\prime}\right) \mid \bar{\sigma}. \phi(s), \bar{\sigma}. \alpha_{s}(a)) 
\right|,
\end{equation}
where $\bar{\sigma} \in \symk$ a permutation acting on $\mathcal{S}_\slots$, $\mathcal{A}_\slots$, and $\mathcal{S}_\slots \times \mathcal{A}_\slots$.

Since we assume the projection property holds: $\bar{\sigma}.\phi(s) = \phi(\sigma.s),  \bar{\sigma}.\alpha_s(a) = \alpha_{\sigma.s}(\sigma.a)$,
the equation holds: 
\begin{equation}
    \hat{T}_\slots(\bar{\sigma}. \phi\left(s^{\prime}\right) \mid \bar{\sigma}. \phi(s), \bar{\sigma}. \alpha_{s}(a)) = 
\hat{T}_\slots(\phi(\sigma.s') \mid \phi(s), \alpha_{\sigma.s}(\sigma.a))
.
\end{equation}

Thus the quantity can be transformed point-wisely as:
\begin{equation}
\lambda_\slots^{\sigma}
=^{(2)} \left| 
\hat{T}_\slots(\phi\left(s^{\prime}\right) \mid \phi(s), 
\alpha_{s}(a)) 
- 
\hat{T}_\slots(\phi(\sigma.s') \mid \phi(s), \alpha_{\sigma.s}(\sigma.a)
\right|.
\end{equation}

By the definition of homomorphism ($\bar{T}$ is $\hat{T}_\slots$, and $T$ is $\hat{T}_\libset$) for transition function:
\begin{equation}
    \bar{T}\left(\phi\left(s^{\prime}\right) \mid \phi(s), \alpha_{s}(a)\right)  \triangleq \sum_{s^{\prime \prime} \in \phi^{-1}\left(\phi(s^{\prime})\right)} T\left(s^{\prime \prime} \mid s, a\right), \forall s, s^{\prime} \in \mathcal{S}, \forall a \in \mathcal{A},
\end{equation}
we substitute the equality
\begin{equation}
    \lambda_\slots^{\sigma}
=^{(3)} \left| 
\sum_{s^{\prime \prime} \in \phi^{-1}\left(\phi(s^{\prime})\right)} \hat{T}_\libset(s'' \mid s,a)
-
\sum_{s^{\prime \prime} \in \phi^{-1}\left(\phi(s^{\prime})\right)} \hat{T}_\libset(\sigma.s'' \mid \sigma.s, \sigma.a) 
\right|.
\end{equation}

By further manipulating the quantity, we get
\begin{equation}
    \lambda_\slots^{\sigma}
=^{(4)}
\sum_{s^{\prime \prime} \in \phi^{-1}\left(\phi(s^{\prime})\right)}
\left| 
\hat{T}_\libset(s'' \mid s,a)
-
\hat{T}_\libset(\sigma.s'' \mid \sigma.s, \sigma.a) 
\right| 
=^{(5)} C \cdot \lambda^\sigma,
\end{equation}
where the sample equivariance error in the full MDP $\lambda^\sigma \triangleq \left| 
\hat{T}_\libset(s' \mid s,a) - \hat{T}_\libset(\sigma.s' \mid \sigma.s, \sigma.a) 
\right|$ is defined in Definition \ref{def:equivariance-error}, and $|\phi^{-1}\left(\phi(s^{\prime})\right)| = {N\choose K}$ is equal for all $s'$, since every combination can come from ${N\choose K}$ possible scenes, and we denote it as $C$.
Finally, we apply expectation over all $\sigma \in \symn$ and arrive at the Proposition \ref{prop:error}.

\section{Additional results}
\label{sec:additional_results}

\begin{table*}[t]
\centering
\caption{
\minor{Results for all methods on the Shapes environment with $K=5$ and $N=5,10,20,30$ (four numbers in each cell). 'OM' stands for out of GPU memory, where we limit the usage to 10GB.
We report for memory usage on $N=20$.
}
}
\scriptsize
\begin{tabular}{l|rrr|rrrr}\toprule

\textbf{Shapes} & \textbf{MRR} (\%, 1-step) & \textbf{H@1} (\%, 5-step) & \textbf{MRR} (\%, 5-step) & \textbf{Train} (MRR, 5-step) & \textbf{Gap}  (MRR, 5-step)  \\
\midrule
$\Sigma_N\text{-CSWM}$  &\textbf{100, 100, 99.9}, \textit{OM} &\textbf{99,9, 99.8, 99.8}, \textit{OM} & \textbf{99.9, 99.9, 99.9}, \textit{OM} & \textbf{100, 100, 100}, OM & 0.0, 0.0, 0.1, \textit{OM}\\
\midrule
$\Sigma_K\text{-CSWM}$  & 100., 80.4, 70.8, 74.1 & 99.8, 32.7, 22.6, 20.1 & 99.9, 43.7, 32.2, 29.4 & 100., 100., 100., 100. &  0.1, 55.3, 67.8, 70.6 & \\
$\Sigma_K\text{-CSWM(CA)}$  & 95.0, 72.0, 71.4, 80.9 & 77.5, 18.7, 15.0, 15.4 & 85.0, 26.8, 22.7, 24.3 & 96.3, 96.5, 96.1, 98.0 & 11.3, 69.7, 73.4, 73.7   \\
$\text{C-WM(N)}$  & 83.6, 81.4, 25.8, 12.9  & 49.6, 41.0, 4.4, 2.2  & 61.8, 55.0, 11.2, 7.6 & 88.0, 96.6, 41.5, 33.9 & 26.2, 41.6, 30.3, 26.2 \\
$\text{MONet(N)+BM}$ & 12.6, 73.9, 35.9, \textit{OM} & 4.5, 11.5, 44.4, OM& 2.0, 20.2, 55.9, \textit{OM}&  7.0, 64.5, 84.8, \textit{OM} & 5.0, 44.3, 29., \textit{OM} \\
\midrule
$\text{HOWM}$ (ours) & 96.7, 94.3, 98.7, 99.6 &76.8, 51.7, 54.8, 35.7 &84.3, 63.2, 65.3, 47.7 &95.5, 94.3, 95.3, 94.7 &11.2, 31.1, 31.3, 43.5 \\
\bottomrule
\end{tabular}
\vspace{-10pt}
\end{table*}

\begin{table*}[t]
\centering
\caption{
Results for all methods on the Rush Hour environment with $K=5$ and $N=5,10,20$ (three numbers in each cell). 'OM' stands for out of memory for GPU training, where we limit the memory usage to 10GB.
}
\scriptsize
\begin{tabular}{l|rrr|rrrr}\toprule

\textbf{Rush Hour}  & \textbf{MRR} (\%, 1-step) & \textbf{H@1} (\%, 5-step) & \textbf{MRR} (\%, 5-step) & \textbf{Train} (MRR, 5-step) & \textbf{Gap}  (MRR, 5-step)  \\
\midrule
$\Sigma_N\text{-CSWM}$ &\textbf{100, 100, 99.9} &\textbf{100, 100, 99.9} & \textbf{99,9, 99.8, 99.8 } & \textbf{100, 100, 99.9} & \textbf{0.01, 0.02, 0.01}\\
\midrule
$\Sigma_K\text{-CSWM}$ & 100. ,  66.9,  47.6 & 99.8, 18.9,  8.  & 99.9, 29.5, 13.7 & 100. ,  95.1,  99. &  0.01,  66.0,  85.0 \\
$\Sigma_K\text{-CSWM(CA)}$ & 99.2, 55.4, 80.2 &91.8,  8.7,  8.7 & 94.9, 15.5, 15.8 & 99.1, 100. , 100. & 4.2, 84.5, 84.2 \\
$\text{C-WM(N)}$  & 55.5, 67.8, 80.6 & 8.5, 13.4, 36.4 & 23.7, 24.5, 45.3 &72.5, 95.2, 99.5 &  48.8, 70.7, 54.1 \\
\midrule
$\text{HOWM}$ (ours)  & 99.2, 98.5, 99.7 &88.9, 77.6, 66.3 &92.3, 84.2, 75.1  &97.0, 96.9, 98.0 &4.7, 12.7, 22.9 \\
\bottomrule
\end{tabular}
\end{table*}

\textit{(HOWM)}
The results of HOWM show that our soft approach, with Action Attention module, is able to (1) end-to-end learn the approximately correct binding between $K$ slots and $N$ factorized actions and (2) correctly align adjacent time steps for computing contrastive loss, only through transition data $(s, a, s')$ with the differentiable Aligned Loss.
Furthermore, recall the multi-step evaluation, our approach unrolls the model in the latent $K$-slot MDP $\bar{s}_{t+1}=T(\bar{s}_t,\bar{a}_t)$ and aligns with the target states $s^\uparrow_{t+k} \approx s_{t+k}$, which only requires $\symk$-equivariance (thus $O(K^2)$ complexity).
The results demonstrate that this latent approach is able to achieve reasonable results with $\symn$ exact methods for intermediate $N \le 20$ while consumes much less resources, and still has potential to scale to $N=30$ and more.
\minor{However, we found for long-term prediction (more than 5 steps), it is still quite challenging to achieve (1) action binding and (2) compositional generalization for large library perfectly, as $N=30$ for 5-step shows.}
One potential reason is that, the learned representation module (Slot Attention in our case) does not guarantee perfect object representations for long steps. This seems to affect the learning of action binding (if slots are inaccurate, actions cannot be bound correctly) and the long-term evaluation (if some objects missed at some step, all following steps can be wrong).

\textit{(Representation bottleneck)}
In general, the learned object representation seems to be a main bottleneck of the decoupled training. 
For example, if Slot Attention \citep{locatello2020object} cannot reliably separate and segment objects into slots, our downstream Action Attention module would struggle in binding actions with corresponding objects.
Once a good representation module is learned, we found our approach has very small variance in transition learning. 
\minor{Also, the variance seems to highly depend on the number of library objects $N$. We choose the best representation checkpoints to train, but we found $N=30,50$ can succeed in most runs, while $N=10,20$ is likely to miss objects in $1/2$ or $1/3$ of runs.}

\textit{(Training resource)}
We highlight the consumed resources in training $\symn$-equivariant models.
As our framework suggests, these models should provide perfect compositional generalization, if it also has \textit{action binding}. 
However, in practice, $\symn$-equivariant models require more training resources and cannot scale up. In Shapes, both $\text{MONet+BM}$ and $\symn\text{-CSWM}$ can only scale to around $N\approx 20$.
For $\text{MONet+BM}$, we train representation with $N$ object slots and $\symn$ transition model separately, and each of them may take around 10GB for $N=20$.
For large $N$'s, the training time of $\text{MONet+BM}$ is approximately two days using one GPU.
$\symn\text{-CSWM}$ jointly learns $N$ dedicated object masks, corresponding to the ordering of $N$ factorized actions.
We do note that, if there is enough resource available for $N$ dedicated object masks, this approach provides the best available solution.
It does not seem to be an universal approach because of independently training $N$ object masks, where $N$ is the number of \textbf{all} possible library objects that can possibly appear in an environment.

\begin{figure*}[t]
    \vspace{-10pt}
    \subfigure{
        \includegraphics[width=0.40\textwidth]{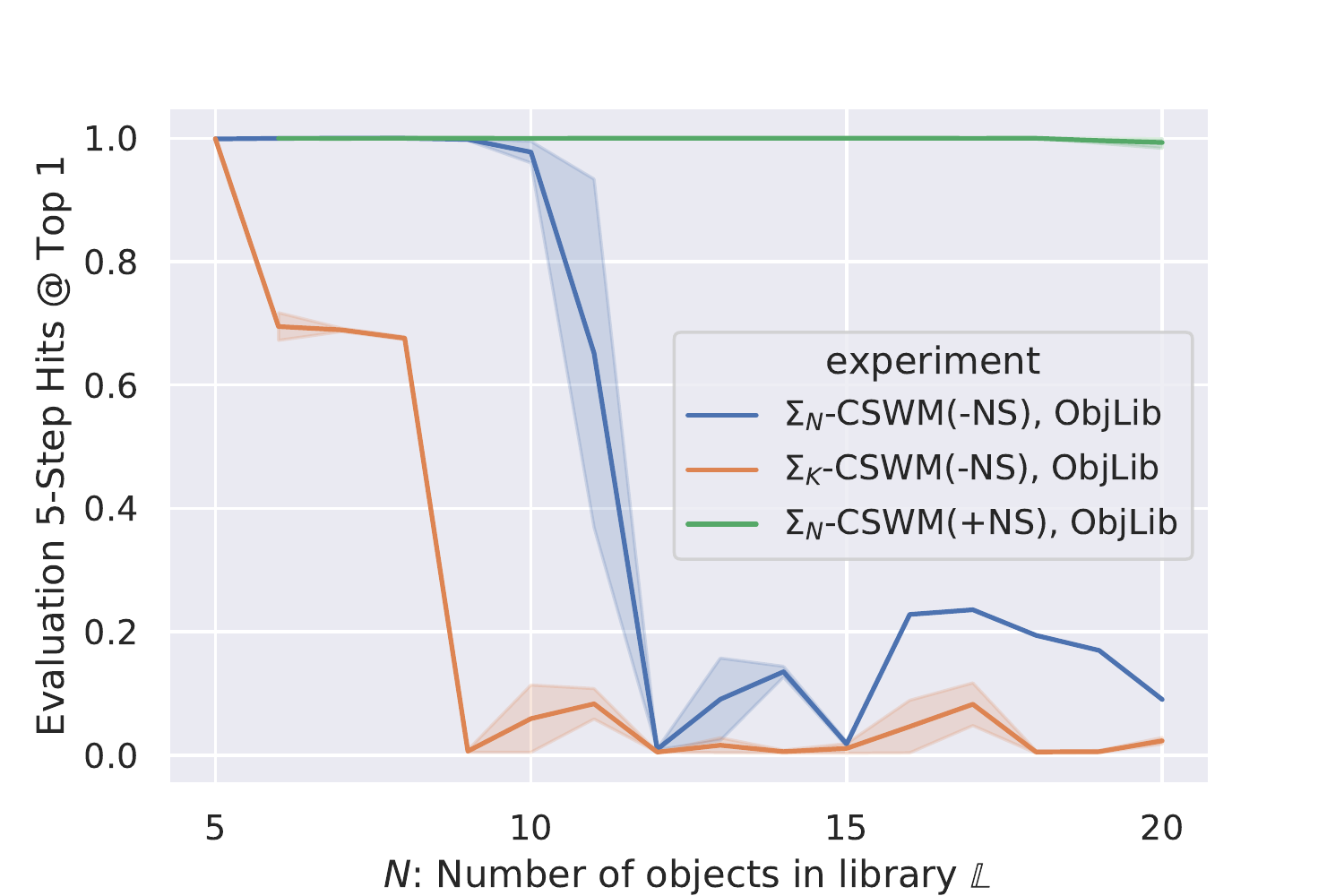}
    } 
    \hfill
    \subfigure{
        \includegraphics[width=0.40\textwidth]{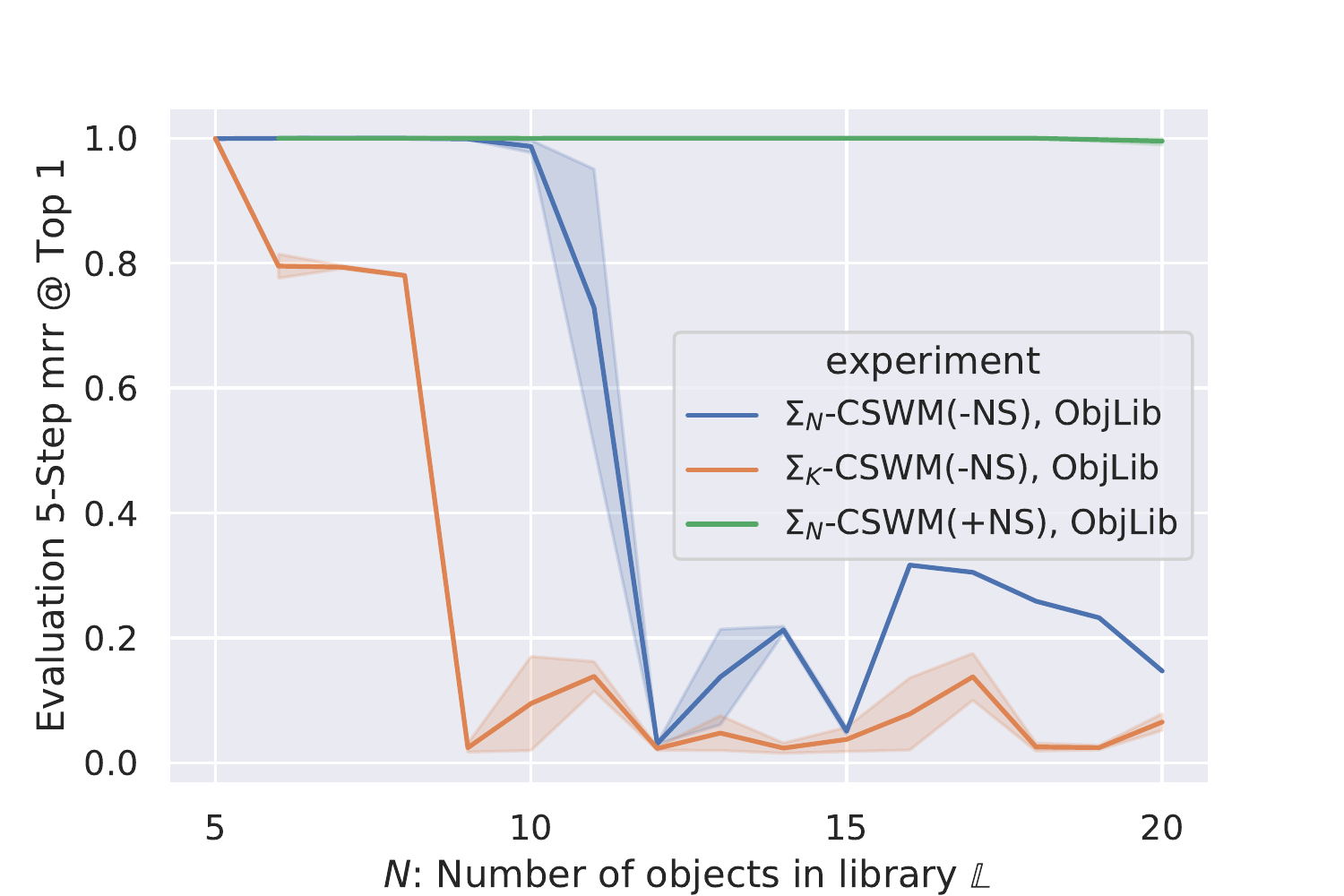} 
    } 
    
    \caption{
        Comparison of different negative sampling strategy. Left figure shows the result of H@1 step 5, and right figure shows the result of MRR step 5.   
    }
    \label{fig:appendix-ns} 
\end{figure*}

\textit{(Comparison between training with controlled and non-controlled negative sample)}
We change the negative sampling strategy. In particular, we select half of negative samples from the same object combinations and the other half from different object combinations. Figure \ref{fig:appendix-ns} shows the impact of the controlled negative sampling strategy. With controlled negative sampling strategy, the performance of $\Sigma_N\text{-}\rm{CSWM}$ is improved by a large extent. We argue that the $\Sigma_N\text{-}\rm{CSWM}$ is inclined to go to local minimum without controlled negative sampling strategy.

\textit{(Negative sampling issue)}
\minor{We note for a sampling issue existing in two perspectives: (1) negative sampling in contrastive loss, (2) sampling reference states in computing Hits and MRR.}
The issue in negative sampling in contrastive loss used in \citep{kipf2019contrastive} is that, since the number of possible scenes $N \choose K$ increases greatly with $N$, if we train using randomly sampled negative states in contrastive loss \citep{kipf2019contrastive}, we found the model will most likely sample negative states from other episodes, thus it may find a degenerate solution that just classifies scenes instead of objects in next states, as observed in \cite{biza2021impact}.
To fix the issue, in all training, we use $50\%$ negative states from same object combinations and the other $50\%$ from different object combinations.
\minor{For the reference states in computing Hits and MRR, it needs to keep enough number of states from each scene to ensure each scene has enough cover, and also enough scenes to ensure diversity.}

\begin{figure}[t]
\centering
    \includegraphics[width=0.9\textwidth]{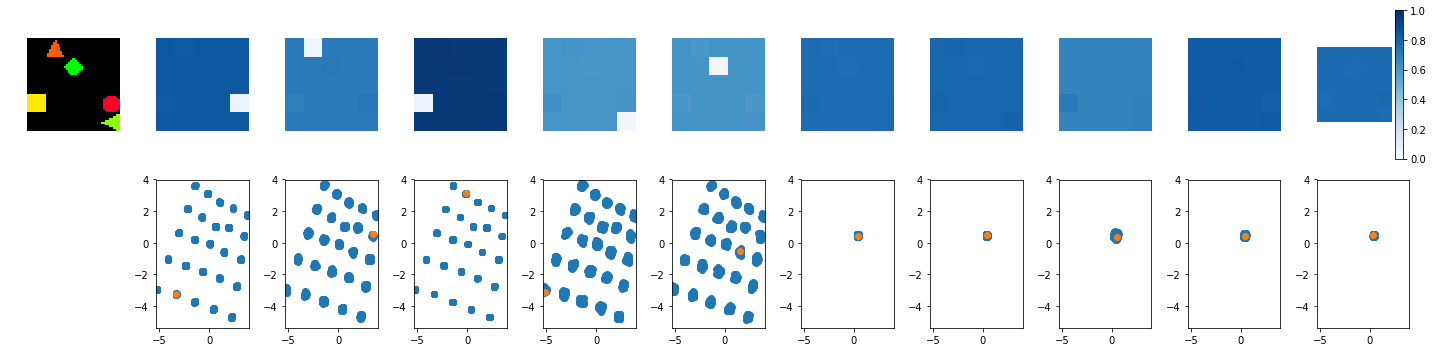}
    \includegraphics[width=0.9\textwidth]{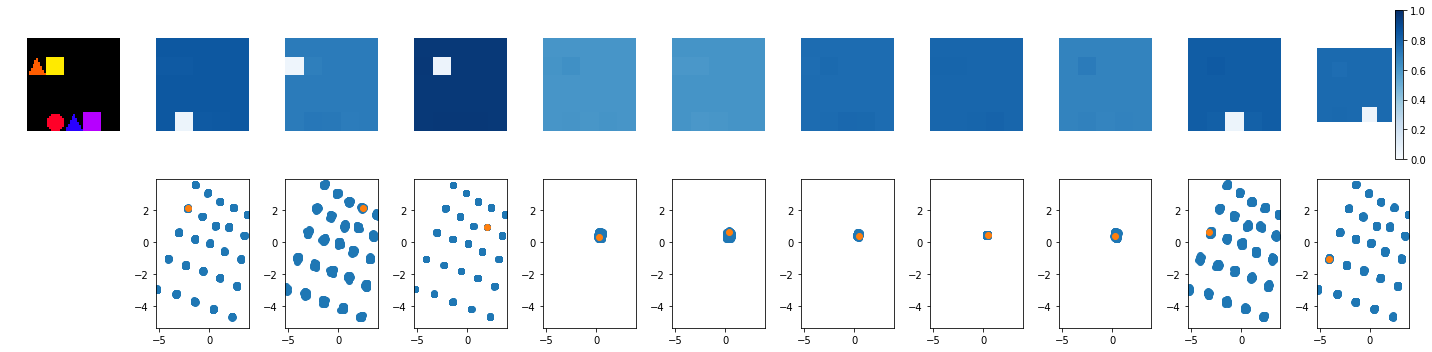}
    \caption{
    \minor{Learned object masks and embeddings in Shapes in two \textit{unseen} evaluation scenes for $\Sigma_N\text{-}\rm{CSWM}$ (Exact CG). Present objects share similar latent structure and the model shows object replacement symmetry. Blue dots are from each individual object.}
    }
    \vspace{-10pt}
    \label{fig:vis1} 
\end{figure}

\section{Additional Experiment Details}
\label{sec:appendix-experiment}

We experiment several approaches that can (2) learn object representations from images and (2) then object-structured world models.
We further discuss their compositional generalization ability in the next section.
\textbf{C-SWM} \cite{kipf2019contrastive} jointly learns object representations and GNN transition model with contrastive loss. 
We experiment two variants: (1) $\Sigma_K\text{-}\rm{CSWM}$ with $K$ nodes in GNN (nubmer of objects on a scene), and (2) $\Sigma_N\text{-}\rm{CSWM}$ with $N$ nodes ($N$ = size of library).
\textbf{MONet} \cite{burgess2019monet} learns object representations using sequential attention and VAE to decompose scenes to multiple slots. 
We implement a $N$-slot version and applies bipartite matching for learning a GNN transition model on top of it, where we name it as $\text{MONet(N)+}\Sigma_N\text{+BM}$.
\textbf{(Resource issue)}
$\text{MONet(N)+}\Sigma_N\text{+BM}$ and $\Sigma_N\text{-}\rm{CSWM}$ consumes CPU and GPU memory intensively, so we can only have small $N$.
We also tried \textbf{STOVE} \cite{kossen2019structured}, which uses RNN with depth of $N$ and requires even more memory, thus cannot finish training for even $N=K=5$.

\subsection{Compositional generalization in the approaches}

We experiment several approaches that can learn object representations from images and a latent object-structured world models.
We classify the approaches based on how well they \textbf{should} achieve compositional generalization ("CG") and analyzing how they achieves CG.

\textbf{(1, Exact CG)}
The first class should achieve exact CG under our framework, which is the $\symn$-equivariant methods with correct action binding.
This only includes $\Sigma_N\text{-}\rm{CSWM}$, which is a variant that outputs $N$ object masks and has $N$ nodes in its transition GNN.
The way it achieves CG relies on learning $N$ \textit{dedicated} object masks, whose order is decided by $N$ factorized actions.
In other words, it \textit{implicitly} bind objects and actions correctly, by fixing the order of object masks. This approach is \textbf{not} universal, since this requires $N$ \textit{dedicated} object masks (slots).

\textbf{(2, Not guaranteed CG)}
If a method misses any of the necessary conditions (in three steps), it cannot achieve (perfect) CG.
We leave (1) object extraction step untouched and focus on (2) action binding and (3) $\symn$-equivariant transition model.
$\text{MONet(N)+BM}$ uses bipartite matching in optimizing transition loss $\texttt{BM}(T(s,a), s')$.
However, since it does \textit{not} have any \textit{action binding} mechanism, the transition network $T(\tilde{s},\tilde{a})$ has no clue to match object slots $\tilde{s}$ and factorized actions $\tilde{a}$.
To break $\symn$-equivariance, we simply replace the $N$-slot GNN and shared encoder with a flat MLP, named $\text{C-WM(N)}$.
We also use two variants of CSWM with $K$ slots, where $\Sigma_K\text{-CSWM}$ receives \textit{factorized} actions only from $K$ scenes, and $\Sigma_K\text{-CSWM(CA)}$ means we copy actions from $N$ objects to all $K$ slots, since no action binding is deployed.

\textbf{(3, Approx. CG)} As comparison, HOWM achieves soft compositional generalization by learning binding objects and actions with \textit{action attention}, which is easier to scale up to large $N$.

\subsection{Implementation and training setup}
\label{sec:additional-baselines}

\textbf{(Data)}
In generating training and evaluation dataset, we guarantee that (1) the combinations of objects in training dataset are different from those of evaluation dataset, and (2) training data contains all $N$ objects in the library. 
\minor{In practice, we found it is critical to keep all individual objects seen in training, or at evaluation the model would very likely mess up objects' colors or shapes, as further detailed in Appendix \ref{sec:compgen}.}
\minor{We use the same object library for all experiments for lower randomness.}
We use 1k episodes in training and $10$ episodes of length $100$ for $100$ scenes, and 10k episodes of length $10$ for evaluation.
Additionally, we make sure that $90\%$ of the transitions have objects moving (filtering collisions), so there is denser signals for relating actions and moved objects.

\textbf{(Training)}
For $\Sigma_K\text{-}\rm{CSWM}$ and $\Sigma_N\text{-}\rm{CSWM}$, we jointly train the entire model \cite{kipf2019contrastive}. 
We train $\text{MONet(N)+BM}$ by first learning the object-oriented representation using pixel reconstruction loss, then freeze the representation module and train the GNN transition model with bipartite matching in the learned latent space.
See more details in Appendix~\ref{sec:appendix-experiment}. %

(\textbf{Metrics})
We measure the (multi-step) dynamics prediction error using two ranking metrics: Hits at Rank 1 (H@1) and Mean Reciprocal Rank (MRR) \cite{kipf2019contrastive}, averaged over 3 runs.
For our approach, we need to align them in the full MDP $\fullmdp$.
\minor{We also report these metrics in training scenes and compute the \textit{generalization gap}, as a proxy for compositional generalization error, because of the lack of object binding information in latent space.}

\begin{itemize}[leftmargin=10pt]
    \item $\Sigma_K\text{-}\rm{CSWM}$ and $\Sigma_K\text{-}\rm{CSWM}\text{-}\rm{CA}$. C-SWMs \cite{kipf2019contrastive} are Contrastively-trained Structured World Models that apply a contrastive approach to learn the representation and the transition model of environments with compositional structure. Specifically, the models learn a set of factorized state variables, which encodes information about each object in the scene. These state variables are then fed into to a graph neural network to model the transitions of the environment. We call this approach \texttt{C-SWM($K$)} because in the original paper, they do not consider scenes with different combinations of object. In other words, they only consider settings with $N=K=5$, where the combination across different episodes remains the same. We first use the model in the original paper as a baseline, which we term $\Sigma_K\text{-}\rm{CSWM}$. Then, we further adapt the model by concatenating the whole action vector to each object representation and then feeding them into the graph neural network, which we term $\Sigma_K\text{-}\rm{CSWM}\text{-}\rm{CA}$. 
    
    \item $\Sigma_N\text{-}\rm{CSWM}$. 
    To further study the capability of C-SWMs, we increase the number of object slots to $N$, i.e., we use $N$ slots to learn the full transition model of the object library. In the original C-SWM model, the number of object slot is equal to the number of objects in the scene, which is $5$ in 2D shapes environment. We also increase the number the action to $4\times N$, where each 4 actions control one specific object.

    \item $\rm{C}\text{-}\rm{WM}(N)$. 
    We use non-factorized encoder MLP and transition MLP as a baseline. The purpose of this baseline is to see how the most naive model will perform in terms of compositional generalization. 
    
    \item $\text{MONet(N)+}\Sigma_N\text{+BM}$. 
    MONet \cite{burgess2019monet} decomposes images into different slot variables, each of which represents information about objects or background. It applies a recurrent attention network to produce attention masks for objects and background, and then feed each mask together with the image to a component VAE to learn object-oriented representations. 
    We divide the experiment into two phrases. First, we train the MONet model using the image input to learn the object-oriented representation. The hyperparameters of the original MONet do not work well on our dataset because models like MONet rely on inductive biases, and often need to be adjusted to new datasets. In particular, the component VAE is too flexible for the dataset so we have to reduce the dimension of the latent variables to $4$. We further change the structure of the attention U-net and the encoder and decoder of the component VAE.     
    
    Second, we freeze the model parameters of MONet and use the learned object-oriented representation as input of the GNN to learn the transition model of the environment. For $\text{MONet(N)+}\Sigma_N\text{+BM}$, we implement a bipartite matching function between the object representation model and the transition model.

    \item \textit{Other related baselines}. Slot attention module proposed in \cite{locatello2020object} is a component that connects the perceptual representations such as feature maps extracted by CNN and the object-based slot representations. It uses iterative attention to enable each slot to compete for explaining part of the perceptual input. Based on the original \texttt{C-SWM} model, we place the slot attention module between the feature maps extracted by CNN and the object-based slot representations, and the other parts of the model remain the same. STOVE \cite{kossen2019structured} uses RNN with depth of $N$, that require much memory, and cannot finish training for $N=5$.   
    
\end{itemize}

\subsection{Training and Evaluation Setup}
\label{subsec:training}

We match the training procedures with C-SWM \cite{kipf2019contrastive} where possible. For \texttt{C-SWM} baselines and our own model, we joint train the representation module and transition module, similar to C-SWM \cite{kipf2019contrastive}. We train the MONet model using the image input to learn the object-oriented representation. We then freeze the model parameters of MONet and use the learned object-oriented representation as input of the GNN to learn the transition model of the environment.  All models are trained on Nvidia GeForce RTX 2080 Ti GPU with 11GB memory.   

\begin{itemize}[leftmargin=10pt]
    \item $\Sigma_K\text{-}\rm{CSWM}$, $\Sigma_N\text{-}\rm{CSWM}$, and $\rm{C}\text{-}\rm{WM}(N)$. We follow the same training and evaluation settings as in \cite{kipf2019contrastive}. In particular, for training dataset, we generate 1000 episodes, with each 100 time steps; for evaluation dataset, we generate 10000 episodes, with each 10 time steps. Models are trained for 100 epochs. We use Adam optimizer with a learning rate of $5\times10^4$ and batch size of $1024$, margin of hinge loss $\gamma = 1.0$, same as the original paper.        

    \item $\text{MONet(N)+}\Sigma_N\text{+BM}$.   
    We divide the experiment into two stages: object-representation learning and transition model learning. The way we generate the training and evaluation dataset is the same as $\Sigma_K\text{-}\rm{CSWM}$. For the first stage, we only use image observations for training. 
    For MONet \cite{burgess2019monet} implementation, we adapt the code from a third-party implementation\footnotemark{}. We follow the training hyperparameters suggested in the original MONet paper. We use the following settings: flags: \texttt{--geco False --pixel\_std1 0.09 --pixel\_std2 0.11 --train\_iter 1000000 --batch\_size 64 --optimiser rmsprop}. For the second stage, we use the same settings as $\Sigma_K\text{-}\rm{CSWM}$.     
    
\end{itemize}

\footnotetext{\url{https://github.com/applied-ai-lab/genesis}}

\subsection{Additional Environment Details} 
\label{sec:additional-env}

Our environment is built upon the 2D shapes environment in \cite{kipf2019contrastive}.
The original environment consists of 5 objects and the actions are discrete. Actions are moving individual object into four cardinal directions, and thus $|\mathcal{A}| = 5 \times 4$ discrete actions in total. %
At each time step, the agent can only take one action, i.e., moving one object to one specific direction. After the agent took an action, the corresponding object would move into the direction by one step unless the target position is occupied by other objects or out of boundary.  
The agent uses a random policy to collect the experience buffer containing a sequence of tuple $\mathbb{B} = \{ o_t, a_t, o_{t+1} \}$, where ${ o_t }$ is the observation at time step ${t }$,  ${ a_t }$ is the action taken, and  ${ o_{t+1} }$ is the next observation. Observations are $3 \times 50 \times 50$ RGB images containing $K$ different objects. Each object occupies $10 \times 10$ pixels and each location of the overall $50 \times 50 $ pixels can only be occupied by at most one object.

\begin{figure}[t]
\centering
    \vspace{-10pt}
    \subfigure{
        \includegraphics[width=0.21\textwidth]{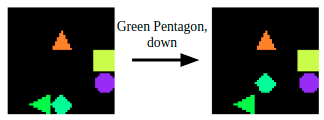} %
    } 
    \subfigure{
        \includegraphics[width=0.2\textwidth]{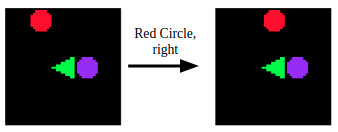}
    }
    \caption{ 
    Example observations of \objlib{}. (\textbf{left}) $K=5$. (\textbf{right}) $K=3$.
    }
    \label{fig:appendix-objlib}
\end{figure}

Based on this environment, we create an object library consisting of a certain number of objects, each of which has a unique shape and color combination. At each episode, we choose $K$ objects from the library and place them to our environment. For training dataset, we randomly sample $K$ objects from the object library except the consecutive sequences. For evaluation dataset, we sample one objects randomly, choose the subsequent consecutive $K$ objects. In this way, we can guarantee that the combinations of objects of training dataset are totally different from those of evaluation dataset. We can also generate dataset with different $Ks$, as shown in Figure \ref{fig:appendix-objlib}. We implement two versions of \objlib{}, named \textbf{Shapes} and \textbf{Rush Hour}.

For \textbf{Rush Hour}, the shapes of objects are chosen from four triangles heading to four different directions, while the colors are all different.
The actions are not move up, down, left and right anymore. Rather, it depends on the heading of the triangle. We define: the heading of the triangle as move forward and the opposite direction as move backward; left of the heading as move left and right of the heading as move right. In other words, there are four different action spaces, each of which corresponds to the triangle heading to one specific direction. For example, if a triangle (car) faces east, then taking action \texttt{forward} results in moving east,  \texttt{backward} west, \texttt{right} south, and \texttt{left} north; if a triangle (car) faces south, then taking action \texttt{forward} results in moving south, etc.

To evaluate compositional generalization of the model, i.e., the capability to understand novel combinations of known objects, we want the combination of objects of the train dataset and evaluation dataset to be as different as possible, while every individual object will appear in both dataset. One analogy of this is to think of it as a chemical experiment \cite{keysers2019measuring}. Each example (compound) is generated by combining primitive elements (atoms). We can view chemical experiments as rearranging atoms to generate new compounds. The atoms are contained in both train and evaluation dataset, while there are unseen compounds in evaluation dataset.

\paragraph{Generating data with controllable object combinations for negative samples.}

We generate data with controllable object combinations for negative samples. In particular, for training dataset, we select negative samples from the same object combination with probability $\epsilon$ and from different object combinations with probability $1-\epsilon$. For training dataset, we only sample some combination of $K$ objects; for evaluation dataset, we sample combinations that are different from each combination in training dataset. For the training dataset, we limit the number of different combinations to $N$, and generate $\# \text{of episodes} / N$ episodes for each combination. During data generation, we save the type of objects (color and shape),  which are then used for all experiments.

\section{Model Architectures} 
\label{sec:appendix-architecture}

We provide the details of model architecture of the methods we use in experiment in several tables.

$\Sigma_N\text{-}\rm{CSWM}$ model architecture is described in below tables. Table 1 describes the object extractor and Table 2 describes the object Encoder. The transition model is GNN-based, where both the node and edge model are the same architecture as object Encoder.

\begin{table}[]
\centering
\caption{Object Extractor for $\Sigma_N\text{-}\rm{CSWM}$ baseline}
\label{tab:my-table}
\begin{tabular}{llll}
\toprule
type                            & size/channel          & Activation            & Comment               \\ \midrule %
Conv 10 x 10                    & 32                    & ReLU                  & Stride: 10            \\ 
BatchNorm2d                     & -                     & -                     & -                      \\  
Conv 1 x 1                      & N                      & Sigmoid               & Stride: 1             \\ \bottomrule %
\end{tabular}
\end{table}

\begin{table}[]
\centering
\caption{Object Encoder for $\Sigma_N\text{-}\rm{CSWM}$ baseline}
\label{tab:my-table}
\begin{tabular}{llll}
\toprule
type                            & size/channel          & Activation            & Comment               \\ \midrule 
Linear & 25 x 512                                    & ReLU                  & -            \\ 
LayerNorm                    & -                     & -                     & -                      \\  
Linear & 512 x 512                                    & ReLU                  & -            \\ 
LayerNorm                    & -                     & -                     & -                      \\  
Linear & 512 x 2                                    & ReLU                  & -            \\ \bottomrule
\end{tabular}
\end{table}

\begin{table}[]
\centering
\caption{Monet attention downsampling}
\label{tab:monet-u-net}
\begin{tabular}{llll}
\toprule
type                            & size/channel          & Activation            & Comment               \\ \midrule %
Conv 3 x 3                    & 32                    & ReLU                  & Stride: 1, Padding: 1            \\ 
BatchNorm2d                     & -                     & -                     & -                      \\  
Conv 3 x 3                    & 32                    & ReLU                  & Stride: 1, Padding: 1             \\  
BatchNorm2d                     & -                     & -                     & -                      \\ 
Conv 3 x 3                    & 64                    & ReLU                  & Stride: 1, Padding: 1             \\
BatchNorm2d                     & -                     & -                     & -                      \\ 
Conv 3 x 3                    & 64                    & ReLU                  & Stride: 1, Padding: 1             \\
BatchNorm2d                     & -                     & -                     & -                      \\ 
Conv 3 x 3                    & 64                    & ReLU                  & Stride: 1, Padding: 1             \\ 
BatchNorm2d                     & -                     & -                     & -                      \\ 
\bottomrule %
\end{tabular}
\end{table}

\begin{table}[]
\centering
\caption{Monet attention upsampling}
\label{tab:monet-u-net-up}
\begin{tabular}{llll}
\toprule
type                            & size/channel          & Activation            & Comment               \\ \midrule %
Conv 3 x 3                    & 64                    & ReLU                  & Stride: 1, Padding: 1            \\ 
BatchNorm2d                     & -                     & -                     & -                      \\  
Conv 3 x 3                    & 64                    & ReLU                  & Stride: 1, Padding: 1             \\  
BatchNorm2d                     & -                     & -                     & -                      \\ 
Conv 3 x 3                    & 32                    & ReLU                  & Stride: 1, Padding: 1             \\
BatchNorm2d                     & -                     & -                     & -                      \\ 
Conv 3 x 3                    & 32                    & ReLU                  & Stride: 1, Padding: 1             \\
BatchNorm2d                     & -                     & -                     & -                      \\ 
Conv 3 x 3                    & 32                    & ReLU                  & Stride: 1, Padding: 1             \\ 
BatchNorm2d                     & -                     & -                     & -                      \\ 
\bottomrule %
\end{tabular}
\end{table}

\begin{table}[]
\centering
\caption{Monet attention baseline}
\label{tab:monet-att-linear}
\begin{tabular}{llll}
\toprule
type                            & size/channel          & Activation            & Comment               \\ \midrule 
Linear 1600 x 128                 &                     & ReLU                  & -            \\ 
Linear 128 x 128                 &                     & ReLU                  & -            \\ 
Linear 128 x 1600                 &                     & ReLU                  & -            \\ \bottomrule
\end{tabular}
\end{table}

\begin{table}[]
\centering
\caption{Monet baseline encoder}
\label{tab:monet-encoder}
\begin{tabular}{llll}
\toprule
type                            & size/channel          & Activation            & Comment               \\ \midrule %
Conv 10 x 10                    & 32                    & ReLU                  & Stride: 10            \\ 
BatchNorm2d                     & -                     & -                     & -                      \\  
Conv 1 x 1                      & N                      & ReLU                 & Stride: 1             \\
Flatten                     & -                     & -                     & -   \\ 
Linear                 &     N x 25 x 512                & ReLU                  & -            \\  
Linear                 &         512 x 512             & ReLU                  & -            \\ 
Linear                 &        512 x 8              & -                 & -            \\ 

\bottomrule %
\end{tabular}
\end{table}

\begin{table}[]
\centering
\caption{Monet baseline decoder}
\label{tab:monet-decoder}
\begin{tabular}{llll}
\toprule
type                            & size/channel          & Activation            & Comment               \\ \midrule %
BroadcastLayer                    & -                    & -                  & -           \\ 
Conv 3 x 3                      & 16                     & ReLU                 & Stride: 1             \\
Conv 3 x 3                      & 16                     & ReLU                 & Stride: 1             \\
Conv 1 x 1                      & 4                     & -                & Stride: 1             \\
\bottomrule %
\end{tabular}
\end{table}

\end{document}